\documentclass[pdflatex, sn-nature, iicol]{sn-jnl}
\usepackage{subfigure}
\usepackage{tabularx}
\usepackage{graphicx}%
\usepackage{multirow}%
\usepackage{amsmath,amssymb,amsfonts}%
\usepackage{amsthm}%
\usepackage{mathrsfs}%
\usepackage[title]{appendix}%
\usepackage{xcolor}%
\usepackage{textcomp}%
\usepackage{manyfoot}%
\usepackage{booktabs}%
\usepackage{algorithm}%
\usepackage{algorithmicx}%
\usepackage{algpseudocode}%
\usepackage{listings}%
\usepackage{ulem}
\usepackage{lmodern}
\usepackage{array}
\usepackage{float}
\usepackage{xr}
\usepackage{afterpage}
\theoremstyle{thmstyleone}%
\theoremstyle{thmstyletwo}%

\theoremstyle{thmstylethree}%

\newcommand{\etal}{\textit{et al}.}

\newcommand{\REV}[1]{{\color{black}#1}}

\begin{document}

\title[Article Title]{Large Language Models for Multi-Robot Systems: A Survey}


\author[]{\fnm{Peihan} \sur{Li}}\email{pl525@drexel.edu}
\equalcont{These authors contributed equally to this work.}

\author[]{\fnm{Zijian} \sur{An}}\email{za382@drexel.edu}
\equalcont{These authors contributed equally to this work.}

\author[]{\fnm{Shams} \sur{Abrar}}\email{sa3868@drexel.edu}

\author*[]{\fnm{Lifeng} \sur{Zhou$^*$}}\email{lz457@drexel.edu}

\affil[]{\orgdiv{Department of Electrical and Computer Engineering}, \orgname{Drexel University}, \orgaddress{\street{3141 Chestnut Street}, \city{Philadelphia}, \postcode{19104}, \state{PA}, \country{USA}}}

\abstract{The rapid advancement of Large Language Models (LLMs) has opened new possibilities in Multi-Robot Systems (MRS), enabling enhanced communication, task allocation and planning, and human-robot interaction. Unlike traditional single-robot and multi-agent systems, MRS poses unique challenges, including coordination, scalability, and real-world adaptability. This survey provides the first dedicated review of LLM integration into MRS. It systematically categorizes their applications across high-level task allocation, mid-level motion planning, low-level action generation, and human intervention. We highlight key applications in diverse domains, such as household robotics, construction, formation control, target tracking, and robot games, showcasing the versatility and transformative potential of LLMs in MRS.
Furthermore, we examine the challenges that limit adapting LLMs to MRS, including mathematical reasoning limitations, hallucination, latency issues, and the need for robust benchmarking systems. Finally, we outline opportunities for future research, emphasizing advancements in fine-tuning, reasoning techniques, and task-specific models. This survey aims to guide researchers in the intelligence and real-world deployment of MRS powered by LLMs. Given the rapidly evolving nature of research in the field, we continuously update the paper list in the open-source \href{https://github.com/Zhourobotics/LLM-MRS-survey}{GitHub repository}.}

\declare{Competing Interests: The authors declare no competing interests.}


\maketitle

\section{Introduction}\label{sec:intro}

The rapid advancement of Large Language Models (LLMs) has significantly impacted various fields, including natural language processing and robotics. Initially designed for text generation and completion tasks, LLMs have evolved to demonstrate problem-understanding and problem-solving capabilities~\cite{zhao2023survey, wei2022emergent}. This evolution is particularly vital for enhancing robot intelligence by enabling robots to process information and make decisions on coordination and behavior accordingly~\cite{kim_survey_2024, jeong2024survey}. With these capabilities, robots can more effectively interpret complex instructions, interact with humans, collaborate with robotic teammates, and adapt to dynamic environments~\cite{wang2024large}. As robotic systems evolve toward more sophisticated applications, integrating LLMs has become a transformative step, bridging the gap between LLM reasoning and real-world robotic tasks.

\begin{figure*}[ht] \label{fig:sec-1-llm-mrs}
    \centering
    \includegraphics[width=1\linewidth]{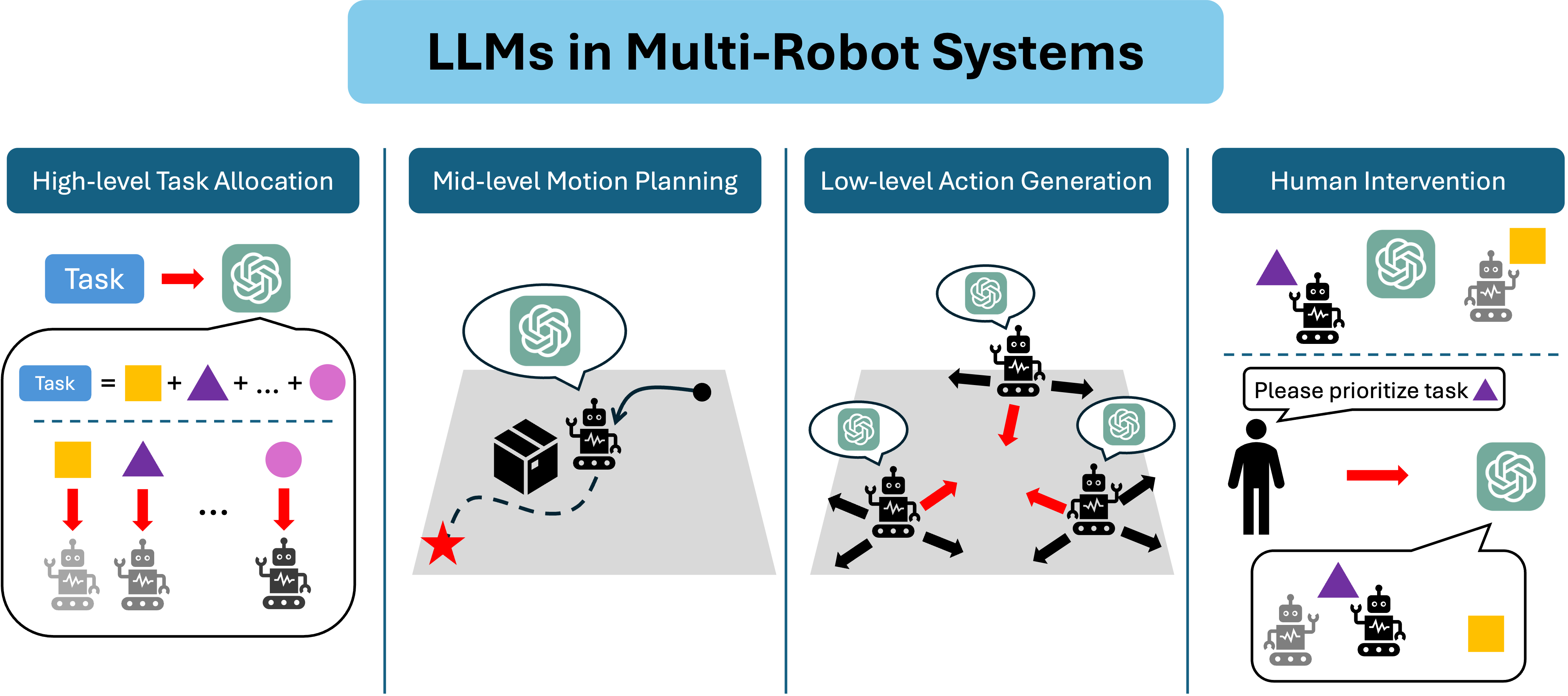}
    \caption{Overview of the applications of LLMs in MRS as introduced in Sec.~\ref{sec:4-LLM-MRS}.}
\end{figure*}

On the other hand, Multi-Robot Systems (MRS), which consist of multiple autonomous robots working collaboratively \cite{queralta2020collaborative, baxter2007multi}, have shown great potential in applications such as environmental monitoring~\cite{ma2018multi, espina2011multi, tiwari2019multi}, warehouse automation~\cite{li2020mechanism, tsang2018novel, rosenfeld2016human}, and large-scale exploration~\cite{burgard2005coordinated, gao2022meeting}. Unlike single-robot systems, MRS leverages collective intelligence to achieve high scalability, resilience, and efficiency~\cite{queralta2020collaborative}. The distributed nature of tasks across multiple robots allows these systems to be cost-effective by relying on simpler, specialized robots instead of a single highly versatile one. Moreover, MRS provides increased robustness, as the redundancy and adaptability of the collective can often mitigate the failures of individual robots \cite{zhou2018resilient, liu2021distributed}. These features make MRS indispensable in scenarios where the scale, complexity, or risk is beyond the capabilities of a single robot.

Despite their importance, MRS introduces unique challenges, such as ensuring robot communication, maintaining coordination in dynamic and uncertain environments, and making collective decisions that adapt to real-time conditions~\cite{gielis2022critical, an2023multi}. Researchers are working to integrate LLMs into MRS to address the unique challenges associated with deploying and coordinating MRS~\cite{chen_scalable_2024, mandi_roco_2024}. For example, effective communication is essential for the MRS to share knowledge, coordinate tasks, and maintain cohesion in the dynamic environment among individual robots~\cite{gielis2022critical}. LLMs can provide a natural language interface for inter-robot communication, allowing robots to exchange high-level information more intuitively and efficiently instead of relying on predefined communication structures and protocols~\cite{mandi_roco_2024}. Furthermore, the problem-understanding and problem-solving abilities of LLM can enhance the adaptability of MRS when given a particular goal without specific instructions. The LLMs can understand the mission, divide it into sub-tasks, and assign them to individual robots within the team based on their capabilities~\cite{liu_coherent_2024, chen_emos_2024}. The generalization ability across different contexts of LLMs can also allow MRS to adapt to new scenarios without extensive reprogramming, making them highly flexible during the deployment~\cite{wang_dart-llm_2024, yu_co-navgpt_2023}.

The application of LLMs in MRS also aligns with the growing need for human-robot collaboration~\cite{hunt_survey_2024}. As operators often do not have expertise in robot systems, using LLMs as a shared interface can enable operators to use natural language to communicate and command the robots to make decisions and complete complex real-world missions~\cite{ahn_vader_2024}. These capabilities enhance the efficiency of MRS and broaden their applicability to domains requiring close human-robot collaboration.

Our paper is inspired by the survey~\cite{guo_large_2024} that comprehensively reviewed LLMs for multi-agent systems (MAS) where abstract agents primarily serve virtual roles. MAS differ from MRS in that the former emphasizes the roles of the agents, while the latter focuses on the interactions between the robots and the physical world. The limited coverage we find regarding MRS in their work pertains to LLM-embodied agents, but it still skims over related work and lacks detailed summaries. Hence, we recognize the necessity of summarizing recent works on using LLMs in MRS for task allocation and planning, motion planning, action generation, and human intervention. Fig.~\ref{fig:sec-1-llm-mrs} illustrates the four categories outlined in this survey paper. We hope this survey can assist researchers in understanding the current progress of using LLMs in MRS, the challenges we face, and the potential opportunities to enhance multi-robot collective intelligence.

We structure our survey paper as follows to better provide a comprehensive introduction to researchers interested in applying LLMs to MRS. Sec.~\ref{sec:2-back} lays the background for the MRS and LLMs for individuals to understand the topics better. Also, we summarize and compare several other existing survey papers about applying LLMs in robotics systems and MAS in general and explain the necessity of our work on MRS. Then, Sec.~\ref{sec:3-comm} reviews the communication structure among the LLMs in the MRS. After that, we review the usage of LLMs at four levels: (1) high-level task allocation and planning, (2) mid-level motion planning, (3) low-level action generation, and (4) human intervention in Sec.~\ref{sec:4-LLM-MRS}. Following reviewing the usage of LLMs, we review based on the applications of the MRS embodied by the LLMs in the real world in Sec.~\ref{sec:5-application}. In Sec.~\ref{sec:6-benchmark}, we summarize the existing benchmark standards for evaluating the performance of LLMs in MRS and the existing simulation environments. In Sec.~\ref{sec:7-discussion}, we identify the challenges and limitations we face and the opportunities and future directions to enhance the LLMs' ability to handle MRS coordination and decision-making, \REV{including a comparative analysis of deployment-relevant factors such as latency, replanning, and communication across representative systems, and a focused discussion of the sim-to-real gap in LLM-based MRS}. Finally, we conclude our paper in Sec.~\ref{sec:8-conclusion}.

\section{Background}\label{sec:2-back}
This section provides background knowledge on MRS and LLMs. While several other research papers have discussed the applications of LLMs in robotic systems, they do not specifically focus on the MRS. We will summarize their contributions and discuss why our survey on facilitating LLMs with MRS is necessary and impactful.

\subsection{Multi-Robot Systems}

An MRS consists of multiple robots that collaborate to complete specific tasks. Unlike single-robot systems, MRS leverages the combined capabilities of multiple robots to perform complex tasks more efficiently, reliably, and flexibly~\cite{alonso2016distributed, rizk2019cooperative,zhou2021multi}. These systems are commonly employed in applications such as search and rescue~\cite{baxter2007multi,luo2011multi,kumar2012opportunities,queralta2020collaborative}, target tracking~\cite{zhou2018active,zhou2019sensor,zahroof2023multi,li2023assignment,liu2024multi}, environmental monitoring~\cite{schwager2011eyes, grocholsky2006cooperative}, coverage and exploration~\cite{burgard2005coordinated,shi2021communication,sharma2023d2coplan,liu2023active,cai2024energy}, and warehouse automation~\cite{alonso2015local, wurman2008coordinating}, where the task's scale or complexity exceeds a single robot's capabilities. When all robots in the team are identical and share the same functionality, the team is called a \textbf{homogeneous} multi-robot team. In contrast, a \textbf{heterogeneous} multi-robot team consists of different types of robots~\cite{parker2008handbook,sharma2020risk,cai2024energy}.
The advantages of MRS include enhanced scalability, as tasks can be distributed among robots~\cite{liu2022decentralized,zhou2022graph,chen2024learning}, and increased resilience, as the failure of one robot can often be mitigated by the others~\cite{zhou2018resilient,zhou2018approximation,ramachandran2020resilient,liu2021distributed,mayya2022adaptive,zhou2022risk,zhou2022distributed,zhou2023robust,shi2023robust,li2024resilient,li2025failure}. In contrast to designing a single, highly versatile robot, MRS usually relies on more uncomplicated, task-specific robots, reducing the cost and complexity of individual units while benefiting from collective intelligence~\cite{jones2004principled}. However, these systems also present unique challenges, particularly in communication, coordination, and decision-making, as robots must operate cohesively in dynamic and uncertain environments~\cite{queralta2020collaborative}. 
Two primary coordination paradigms are commonly employed to manage the interaction and task distribution within an MRS: centralized and decentralized coordination~\cite{yan2013survey, cortes2017coordinated}. In a \textbf{centralized} paradigm, a single coordinator receives all the information and assigns tasks to all robots in the system, allowing for optimized coordination and global task planning. However, centralized systems can become a bottleneck when the group size increases and are vulnerable to single points of failure~\cite{luna2011efficient}. On the other hand, a \textbf{decentralized} paradigm distributes task-level decision-making among the robots, enabling them to operate resiliently~\cite{rizk2019cooperative}. This approach enhances scalability and resilience but often introduces additional complexity to ensure seamless communication and coordination between robots. The choice between centralized and decentralized control depends on the specific application requirements, environmental conditions, and the desired balance between efficiency and robustness~\cite{yan2013survey}.

\subsection{Large Language Models}
LLMs are deep learning models with millions to billions of parameters~\cite{zeng_large_2023}. Initially, the application of the LLMs is for text completion based on the context or text generation from the user's instruction~\cite{zhao2023survey}. LLMs are trained using an extensive collection of text from books, articles, websites, and other written sources. During this training process, LLMs learn to predict the next word in a sentence or fill in missing information using the \textbf{attention} mechanism~\cite{vaswani2017attention}. This pre-training phase enables LLMs to develop a broad understanding of language, grammar, factual knowledge, and reasoning skills~\cite{naveed2023comprehensive}. 

\subsubsection{Fine-tuning and RAG}
While LLMs are pre-trained on a diverse dataset for general tasks, the performance in specialized cases can be suboptimal since the training dataset might not fully cover the special usages~\cite{ding2023parameter, ziegler2019fine}. People can prepare a dataset dedicated to the specialized tasks and retrain the model. However, retraining the entire model is often challenging due to the limited computing resources and numerous parameters within the model~\cite{ding2023parameter}. One solution to address this issue is to use techniques like low-rank adaptation (LoRA) to \textbf{fine-tune} the LLMs with limited computational resources~\cite{hu_lora_2021}. LoRA freezes the pre-trained model weights and injects trainable rank decomposition matrices into each layer of the Transformer architecture~\cite{vaswani2017attention}, significantly reducing the number of trainable parameters for the downstream tasks.

On the other hand, retrieval-augmented generation (\textbf{RAG}) is an alternative technique that integrates external knowledge sources to increase the zero-shot accuracy of the LLMs on specialized tasks~\cite{lewis2020retrieval, huang2023survey}. RAG addresses a key limitation of LLMs' reliance on pre-trained, static knowledge, which may not include domain-specific or up-to-date information. By combining a retrieval mechanism with the generative capabilities of LLMs, RAG allows the model to query external databases or knowledge repositories to retrieve relevant information during runtime~\cite{gao2023retrieval}. This retrieved data is then used to guide the model's response, enhancing its accuracy and applicability in specialized contexts. For instance, RAG can provide real-time access to task-specific knowledge or environmental updates for robots, enabling better decision-making in dynamic scenarios~\cite{zhu2024retrieval}. Although RAG introduces additional complexity, such as managing retrieval latency and ensuring data relevance, it offers a powerful method for bridging the gap between static pre-trained knowledge and the dynamic requirements of real-world applications.

\subsubsection{Multimodal LLMs}
Traditional LLMs process only text, limiting their ability to directly interpret sensory data. In multi-robot systems, however, perception is often the primary source of situational awareness, coming from heterogeneous sensors such as onboard cameras, LiDAR, or aerial imagery. \textbf{Multimodal LLMs} address this gap by integrating inputs such as images, video, audio, or structured sensor data into a shared semantic space alongside language~\cite{yin2023survey}. This allows robots to ground natural language reasoning in real-world observations, for example, identifying mission-relevant objects, understanding spatial relationships, or updating plans based on live visual feedback~\cite{kim_survey_2024, wang2024large_survey}. 

Recent advances have led to the emergence of \textit{Vision-Language Models} (VLMs), which couple perception and reasoning, and \textit{Vision-Language-Action Models} (VLAs), which go further by linking perception and reasoning directly to executable actions. These models extend the role of LLMs in MRS from purely symbolic reasoning to perception-grounded decision-making and, in the case of VLAs, to closed-loop perception-action execution~\cite{zhong2025survey}. As discussed in Sec.~\ref{sec:vlm-vla}, this evolution opens new possibilities for coordinated, perception-driven, and time-critical decision-making in multi-robot teams.

\subsection{Related Survey Papers}
Several survey papers have reviewed applications of LLMs in the robotics and multi-agent field.
Firoozi \etal~\cite{firoozi_foundation_2023}, Zeng \etal~\cite{zeng_large_2023}, and Kim \etal~\cite{kim_survey_2024} all explored how LLMs and foundation models could enhance robotics in areas like perception, decision-making, and control. While they share this focus, their approaches and scopes differ. Firoozi \etal~\cite{firoozi_foundation_2023} provided a broad overview of foundation models in robotics, emphasizing their adaptability across various tasks but without specific attention to MRS. Zeng \etal~\cite{zeng_large_2023} focused on the applications of LLMs in robotics, categorizing their impact on single-robot systems in areas like control and interaction without exploring collaborative systems. Wang \etal~\cite{wang2024large_survey} concentrated on summarizing the applications of LLMs for manipulation tasks for a single robot. Kim \etal~\cite{kim_survey_2024} divided LLM applications into communication, perception, planning, and control, offering practical guidelines for integration, but their work is also centered on single-robot applications. Hunt \etal~\cite{hunt_survey_2024} explored the use of language-based communication in robotics, categorizing applications of LLMs based on their roles in robotic systems, such as tasking robots, inter-robot communication, and human-robot interaction. Their focus is primarily on language as a medium for interaction without addressing the unique challenges of MRS. Guo \etal~\cite{guo_large_2024} reviewed LLM-based MAS, exploring their applications in problem-solving and world simulation. Although their work included embodied agents, their emphasis is on general multi-agent frameworks, which focus on abstract roles and interactions within systems that may not require physical embodiment or real-world interaction. Kawaharazuka \etal~\cite{kawaharazuka_real-world_2024} examined real-world applications of foundation models in robotics, focusing on replacing components within robotic systems but without addressing inter-robot collaboration or the collective intelligence of MRS.

None of these surveys addresses the challenges and opportunities of integrating LLMs into MRS. While MAS provide a generalized framework for understanding roles and interactions, they are often abstract and virtual, lacking the physical embodiment and real-world constraints that characterize MRS~\cite{guo_large_2024}. MRS requires actual physical robots to collectively perceive, decide, and act within dynamic and uncertain environments, posing unique challenges in communication, coordination, and decision-making that go beyond the scope of virtual agents~\cite{wallkotter2021explainable}. Moreover, MRS uniquely benefits from improved scalability, failure resilience, and cost-effective collective operations, making them fundamentally different from single-robot systems or general multi-agent frameworks. This gap highlights the need for a dedicated survey that explores how LLMs can facilitate communication, coordination, and collaborative task execution in MRS, providing critical insights into this emerging and impactful area of research.

\REV{Beyond this high-level observation, several concrete assumptions common in LLM-based MAS no longer hold when agents are embodied. First, MAS benchmarks such as Werewolf, Avalon, and software-development sandboxes generally treat actions as atomic symbolic events that always succeed~\cite{guo_large_2024}, whereas an embodied action must respect continuous-time dynamics and the simultaneous actions of other robots. Chen \etal~\cite{chen_why_2024} show that GPT-4 achieves only 20\% success on eight-agent multi-agent path finding (MAPF) in a room map and 0\% in a maze, despite performing well in empty-grid MAS settings. Second, MAS benchmarks typically assume perfect observability. RoCo~\cite{mandi_roco_2024} reports that real-world task performance is primarily bottlenecked by incorrect object detection from the onboard vision pipeline, since the dialog layer then operates on noisy object estimates rather than ground-truth state. Third, MAS benchmarks commonly assume lossless, turn-synchronous message passing and negligible deliberation latency, both of which break under field conditions. Section~\ref{sec:7-discussion} discusses these deployment constraints in detail. Here we note their direct effect on transferred MAS techniques, for example Li \etal~\cite{li_challenges_2024}, where round-based LLM planning cannot satisfy the velocity-alignment requirement of continuous-time flocking and five-agent formations collapse toward the group centroid. Finally, MAS agents are often role-profiled but capability-symmetric, whereas MRS agents have asymmetric observations and skills, so a directive such as ``pick up the bread'' may be physically infeasible for some recipients~\cite{mandi_roco_2024}.}

\REV{These breakdowns have measurable effects on transferred performance. Fully decentralized LLM dialog, which works well in MAS software-development and social-deduction benchmarks, performs poorly in the multi-robot grid environment of Chen \etal~\cite{chen_scalable_2024} as the team grows beyond a few agents, which the authors attribute to peer-to-peer dialog history exhausting the context token budget. One-shot chain-of-thought prompting, a common technique in MAS reasoning, also fails on the MAPF maze setting reported by Chen \etal~\cite{chen_why_2024}, and hybrid architectures with external syntactic checkers or centralized priming are needed to recover scalability. These results indicate that reported MAS success rates on token-level benchmarks are poor predictors of MRS performance, and that transferring MAS techniques to robots without accounting for continuous physics, imperfect perception, and lossy communication risks systematically overstating achievable capabilities. Section~\ref{sec:7-discussion} (Table~\ref{tab:deployment}) provides concrete evidence of how current LLM-based MRS handle, or fail to handle, these deployment factors.}
\section{Communication Types for LLMs in Multi-robot Systems}\label{sec:3-comm}

LLMs demonstrate remarkable abilities in understanding and reasoning over complex information. However, their performance can significantly vary depending on the communication architecture employed~\cite{chen_scalable_2024, liu_bolaa_2023}. This variability becomes particularly pronounced in scenarios involving embodied agents, where each agent operates with its own LLM for autonomous decision-making. The independence of these LLMs introduces unique challenges in maintaining consistency, coordination, and efficiency across the MRS. Understanding these dynamics is critical to optimizing LLM-based communication and decision-making frameworks in MRS.

\begin{figure*}[htbp]
    \centering
    \includegraphics[width=1\linewidth]{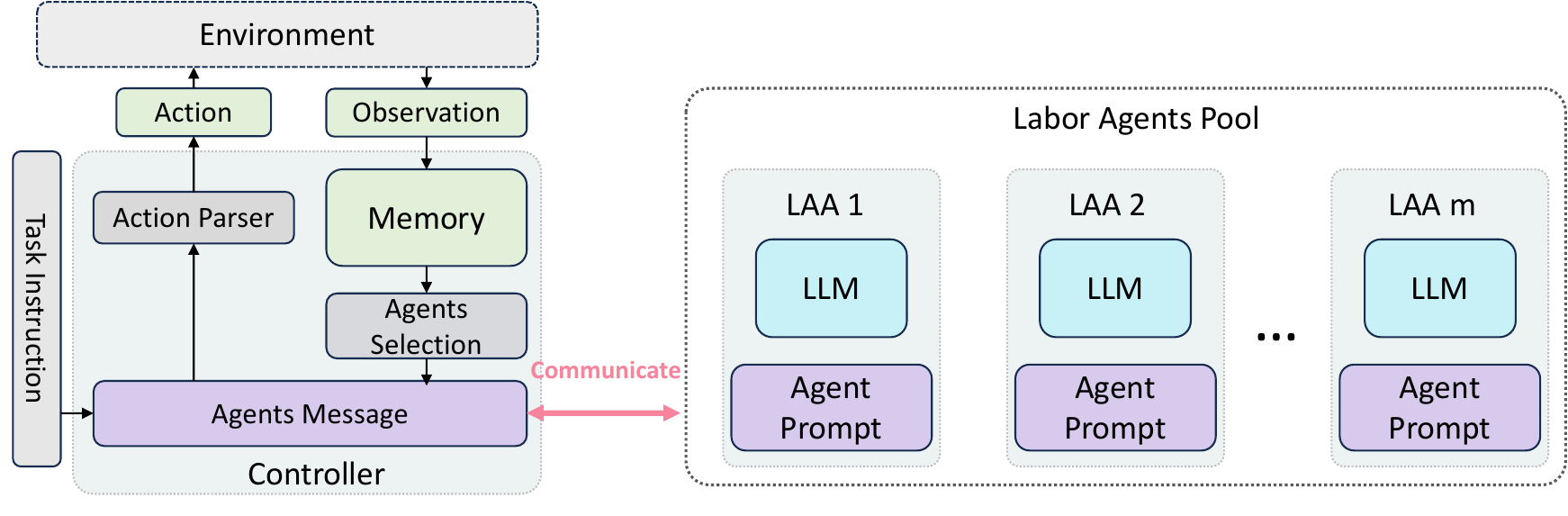}
    \caption{The BOLAA architecture, which employs a controller to orchestrate multiple LAAs~\cite{liu_bolaa_2023}.}
    \label{fig:bolaa_sec3}
\end{figure*}

Liu \etal~\cite{liu_bolaa_2023} provided a comprehensive comparison of LLM-augmented Autonomous Agents (LAAs), analyzing the architectures employed to integrate LLMs into agents. While their work primarily focuses on MAS rather than exclusively MRS, their insights into LLM architectures and agent orchestration offer valuable inspiration for multi-robot applications. Their study begins with a basic structure where LLMs perform zero-shot inference based solely on task instructions and observations. This architecture is then enhanced with a self-thinking loop, incorporating previous actions and observations into subsequent decision-making rounds to improve contextual consistency. They extended the architecture by incorporating few-shot prompts, including example actions to enhance the LLMs' ability to generate effective decisions.
Regarding multi-agent orchestration, Liu \etal~proposed a centralized architecture featuring a message distributor, which relays information to individual agents equipped with their own LLMs. These agents independently process the distributed messages to generate actions, as illustrated in Fig.~\ref{fig:bolaa_sec3}. As discussed in Sec.~\ref{sec:4-LLM-MRS}, several studies have adopted similar self-thinking strategies to improve the consistency and reliability of decisions made by LLMs, demonstrating the utility of this approach in collaborative systems.

\begin{figure}[h]
    \centering
    \includegraphics[width=1\linewidth]{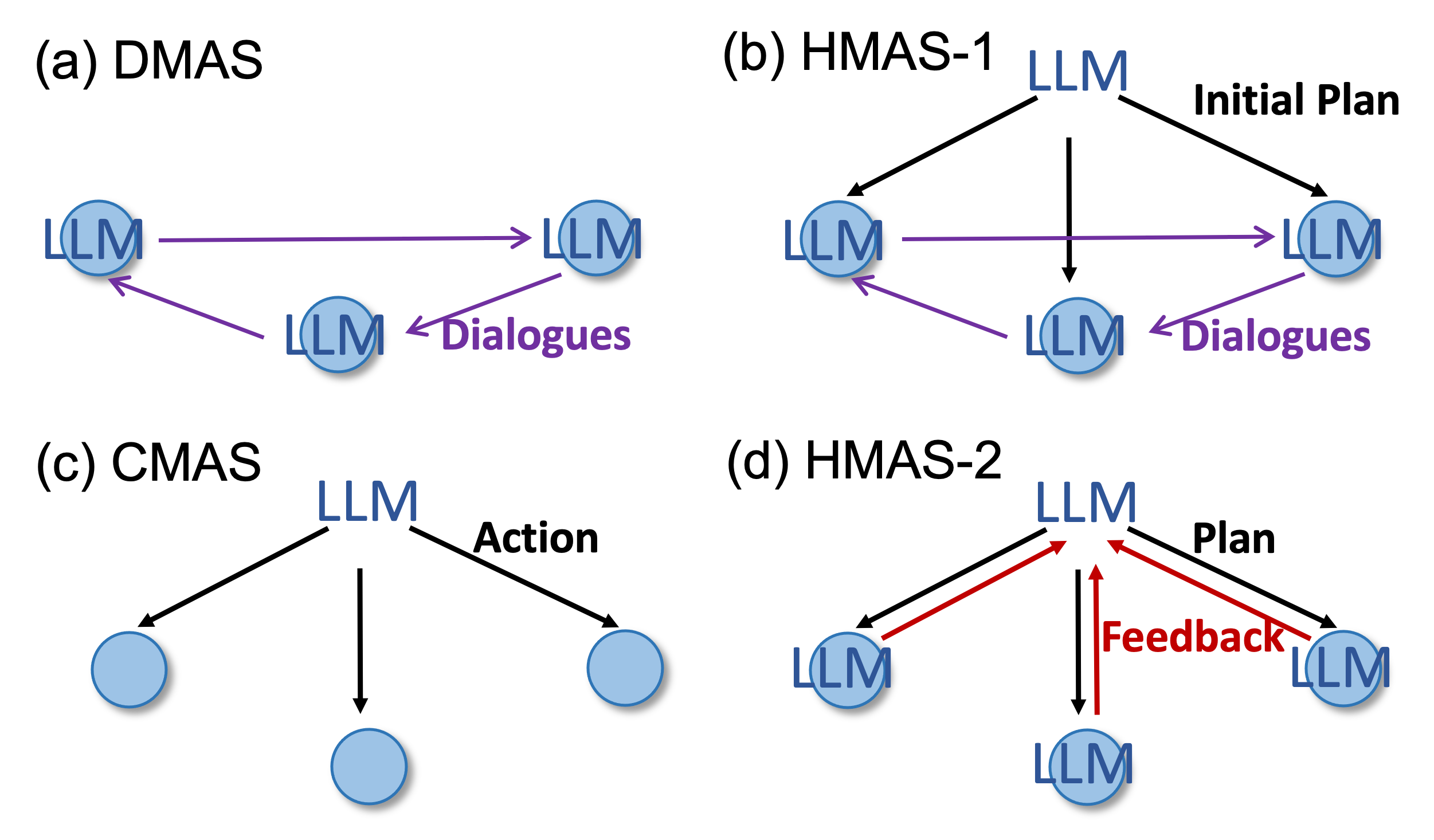}
    \caption{Four LLM-based multi-agent communication architectures introduced in Chen \etal~\cite{chen_scalable_2024}. The circles represent robots that may have actions in the current step and the `LLM' text represents each LLM agent. The overlap between one circle and one `LLM' text means that the robot is paired with one LLM agent to express its special opinions to other agents. The `LLM' text without the overlapped circle represents a central task-planning agent.}
    \label{fig:scalable_sec3}
\end{figure}

Additionally, Chen \etal~\cite{chen_scalable_2024} proposed four communication architectures: a fully decentralized framework (DMAS), a fully centralized framework (CMAS), and two hybrid frameworks that combine the decentralized and centralized frameworks (HMAS-1 and HMAS-2). These frameworks are visually represented in Fig.~\ref{fig:scalable_sec3}. Their study evaluated the performance of these structures in warehouse-related tasks, revealing notable distinctions among them. For scenarios involving six or fewer agents, both CMAS and HMAS-2 demonstrated comparable performance, although CMAS required more steps to complete tasks. In contrast, the performance of DMAS and HMAS-1 was notably inferior. Furthermore, their experiments indicated that HMAS-2 outperformed CMAS in handling more complex tasks, suggesting that hybrid frameworks with optimized structures offer greater scalability and adaptability for intricate multi-robot operations.

\section{LLMs for Multi-robot Systems}\label{sec:4-LLM-MRS}
In this section, we categorize the applications of LLMs in MRS into high-level task allocation and planning, mid-level motion planning, low-level action generation, and human intervention scenarios. High-level task allocation and planning involves tasks that demand a higher degree of intelligence among multiple robots, where the LLM is required to exhibit logical reasoning and decision-making capabilities. Mid-level motion planning refers to navigation or path-planning scenarios. Low-level action generation uses LLMs to produce actuator-level commands that directly control robots' posture or low-level motion. On the other hand, human intervention involves using LLMs to interact with human operators and guide task allocation, planning, and execution across all levels. Table~\ref{tab:table1} shows the list of papers based on those four categories.

\REV{To avoid ambiguity, we use the following terms consistently throughout the survey. \textbf{Task planning} and \textbf{task allocation} refer to the high-level decision of what each robot should do. \textbf{Motion planning} refers to generating collision-free paths or navigation trajectories at the mid-level. \textbf{Action generation} and \textbf{low-level control} refer to producing actuator-level commands for the robot hardware. \textbf{Execution} refers to the physical process of carrying out a generated plan or action, regardless of the level. \textbf{Reasoning} refers to the LLM's internal inferential process, which can occur at any level.}
\subsection{High-Level Task Allocation and Planning}
High-level task planning leverages LLMs' advanced reasoning and decision-making capabilities to handle complex and strategic tasks. This scenario often requires allocating tasks among robot teams, developing comprehensive task plans, or solving problems requiring contextual understanding and logic. Here, we explore studies illustrating LLMs' capability in these sophisticated domains.

\begin{figure*}[ht]
    \centering
    \includegraphics[width=1\linewidth]{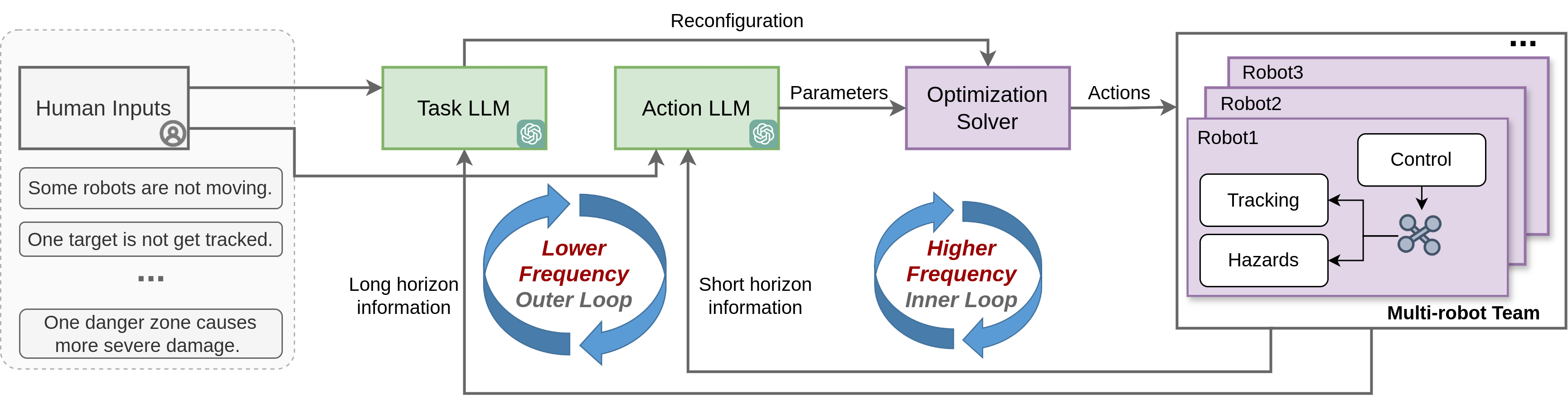}
    \caption{The target tracking architecture proposed by Wu \etal~\cite{wu_hierarchical_2024}. The optimization solver acts as the controller for the multi-robot team. The ``Task LLM" is a high-level task planner, while the ``Action LLM" is a mid-level motion planner integrated with the optimization solver.}
    \label{fig:llm_wu}
\end{figure*}

Recent work has demonstrated that LLMs are capable of allocating tasks among multiple robots. Wu \etal~\cite{wu_hierarchical_2024} proposed a hierarchical LLMs framework consisting of two layers to solve the multi-robot multi-target tracking problem. In this scenario, the LLMs assign targets to each robot for tracking based on the current relative positions, velocities, and other relevant information between the robots and targets. As shown in Fig.~\ref{fig:llm_wu}, the outer task LLM receives human instruction and the long horizon information as inputs to provide strategic guidance and reconfiguration to the robot team. Meanwhile, the inner action LLM takes short horizon information as input and outputs control parameters for the controller. The outputs of the two LLMs are transformed into executable actions through the optimization solver.
In addition, Brienza \etal~\cite{brienza_llcoach_2024} applied VLM and LLM to generate actionable plans for the robotic soccer team. Their approach involved providing the VLM coach with a training set comprising video frames paired with corresponding textual prompts, detailing tasks and constraints. The VLM coach generated schematic descriptions of the video frames along with high-level natural language plans. Two distinct LLMs further refined and synchronized these high-level plans to produce executable strategies suitable for various scenarios. In practical applications, the system selected pre-collected plans based on their similarity to the real-world situation. Additionally, RAG minimizes prompt size and mitigates hallucination, ensuring more reliable outputs. 
Moreover, Lykov \etal~\cite{lykov_llm-mars_2023} developed an MRS to collect and sort colored object sets and count spherical objects. Their approach utilized a fine-tuned LLM to generate Behavior Trees (BTs) for robots to execute tasks and provide feedback to human operators regarding their behaviors. They implemented a single LLM with two LoRA adapters, each handling specific functionalities to enhance efficiency and resource compactness. 
In addition, Ahn \etal~\cite{ahn_vader_2024} introduced an MRS framework featuring a recovery mechanism. The LLM task planner received natural language instructions and a library of low-level robot skills to generate plans for task execution. A key innovation in their system was detecting deviations from the expected task progression and performing error recovery by replanning or seeking assistance from other robots or human operators. 
\REV{A common pattern across these approaches is that each couples an LLM planner with an auxiliary mechanism that compensates for single-LLM brittleness, namely hierarchical decomposition across two LLM layers~\cite{wu_hierarchical_2024}, a VLM-to-LLM refinement pipeline with retrieval augmentation~\cite{brienza_llcoach_2024}, task-specific LoRA adapters~\cite{lykov_llm-mars_2023}, or a dedicated error-recovery channel~\cite{ahn_vader_2024}.} The remaining studies in this domain can be further categorized into two key areas: multi-robot multi-task coordination and complex task decomposition, highlighting the breadth of LLM applications in MRS.

\subsubsection{Multi-Robot Multi-Task Planning} 
In the multi-robot multi-task scenarios, a team of robots is tasked with completing multiple objectives simultaneously. LLMs play a critical role in devising actionable and efficient task distribution strategies in such settings. By interpreting high-level instructions and understanding the context of each task, LLMs can dynamically allocate tasks among robots, ensuring optimal utilization of resources and effective collaboration. This capability enables multi-robot teams to handle complex, multi-faceted operations with increased precision and adaptability. 

Lakhnati \etal~\cite{lakhnati_exploring_2024} proposed a framework where three heterogeneous robots aim to accomplish complex tasks instructed by human operators in VR simulation. First, each robot LLM is given an initial prompt to clarify its role and abilities. A central controller LLM analyzes human descriptions of the task and distributes them to the respective robots. Instructions from human operators can either directly specify what each robot should do (e.g., ``Jupiter needs to move to the dumbbell and pick it up, Neptune and Pluto have to move to the fridge.") or describe the tasks without assigning to specific robots (e.g., ``Three dinner plates have to be put into the trash, and all agents need to end up next to the garbage bin.").
Following this line, Chen \etal~\cite{chen_emos_2024} proposed a centralized framework where a central-planner LLM distributes the human instructions to a multi-robot team. They aim to make a heterogeneous multi-robot team accomplish multiple heterogeneous household tasks. However, the task distribution process they introduced takes the form of a discussion between the ``Central Planner'' LLM and the robot-dedicated agent LLM on each robot. The original task information is a geometric representation from a simultaneous localization and mapping (SLAM) system. It is constructed into a scene context to prompt LLM. The ``Central Planner" LLM first assigns each task to each robot according to its analysis. Then, each robot-dedicated agent LLM provides feedback according to the assigned task, and its robot resume is generated from the robot's URDF code by the robot-dedicated agent LLM. If the task does not match the robot's resume, it prompts the ``Central Planner" for a reassignment. This discussion between LLMs continues until no reassignments are required. 
Lim \etal~\cite{lim2025dynamic} presented a centralized LLM-enabled coordination framework for dynamic task assignment in multi-robot manufacturing systems. The architecture consists of three main components. The Central Controller Agent (CCA) acts as a high-level decision-maker, leveraging LLM reasoning to reassign tasks when disruptions occur. The Robot Agents execute tasks using configuration-defined capabilities and report failures to the CCA. The Sensor Module provides real-time perceptual input to ensure context-aware decisions. When a robot failure is detected, the CCA formulates a structured prompt containing system constraints and objectives, and sends it to the LLM to generate a valid new configuration for an alternative robot. If the LLM-generated configuration violates task constraints, structured feedback is appended, and the process is repeated until success is achieved. This loop ensures that reassignment is both feasible and safe. Once a valid configuration is found, the task is reallocated, and execution resumes seamlessly.
Moreover, Jiang \etal~\cite{jiangexploring} introduced a decentralized swarm robotics framework that leverages LLMs to enable spontaneous inter-robot communication and coordination. Each robot maintains its own LLM session and utilizes natural language to discover peers, share information, and coordinate collaborative actions without predefined roles. The system demonstrates that LLM-driven communication can give rise to emergent social behaviors such as negotiation, assistance, and conflict resolution, pushing swarm robotics toward more human-like, adaptive coordination.
Chen \etal~\cite{chen_scalable_2024} took a step further to investigate the scalability of an LLM-based heterogeneous multi-task planning system. The efficiency and accuracy of four different communication architectures are compared, as shown in Fig.~\ref{fig:scalable_sec3} under four distinct environments, including BoxNet, warehouse, and BoxLift. The results demonstrate that the HMAS-2 structure achieves the highest success rate while CMAS is the most token-efficient. 
On the other hand, Gupte \etal~\cite{gupte_rebel_2024} proposed an LLM-based framework to solve Initial Task Allocation for a multi-robot multi-human system. In this centralized framework, LLM first generates prescriptive rules for each user's objective and then generates experiences based on those rules for each objective. After acquiring a practical knowledge of the rules generated, the LLM's performance is evaluated by inferencing, where the user provides instructions, and the LLM allocates the task according to the rules and experiences. Two distinct RAG workflows are leveraged in the inferencing stage to use the acquired knowledge fully.
Moreover, Huang \etal~\cite{huang2024words} tested the ability of LLMs to solve the multi-robot Traveling Salesman Problem (TSP). By providing appropriate prompts, the LLM plans the optimal paths for multiple robots and generates Python code to control their movements. The study set up three frameworks: single attempt, self-debugging, where the LLM checks whether the generated Python code can be executed, and self-debugging with self-verification, where the LLM checks code executability and verifies whether the execution produces correct results. Their work reveals that LLMs perform poorly in handling such problems, with higher success rates that can only be observed in specific cases, such as the min-max multi-robot TSP. 
Chen \etal~\cite{chen2025multi} presented a multi-agent LLM-based framework for robotic autonomy, integrating task analysis, robot design, and RL path planning. Experiments demonstrate the effectiveness of robot configuration generation and the feasibility of path-planning strategies, showcasing significant improvements in the scalability and flexibility of LLM-driven robotic system development.
Wan \etal~\cite{wan2025toward} introduced a supervised fine-tuned LLM-based planning framework using the novel \textbf{MultiPlan} dataset for embodied multi-robot collaboration. Extensive indoor and outdoor field experiments illustrate significant improvements in task decomposition, robustness, and adaptability to novel environments compared to baseline LLM planners.

\REV{Across the systems reviewed, a common architectural pattern has emerged. Purely centralized LLM coordination offers simplicity but proves unable to self-correct mid-execution and degrades sharply as robot count increases, while purely decentralized approaches incur high communication overhead and often underperform. For example, Chen \etal~\cite{chen_scalable_2024} report that the fully decentralized DMAS achieves only 25\% success on BoxNet1, compared to 82.5\% for the hybrid HMAS-2. The high-performing systems share a hybrid pattern in which a central LLM handles initial allocation while local feedback agents manage validation and re-assignment, as seen in the group discussion followed by parallel execution in EMOS~\cite{chen_emos_2024} and the Central Controller Agent validation loop of Lim \etal~\cite{lim2025dynamic}. A second cross-cutting finding is that closed-loop correction is essential. Open-loop LLM planning consistently underperforms, and most high-performing systems incorporate iterative re-prompting on constraint violations, self-debugging, or RAG-retrieved past experiences~\cite{gupte_rebel_2024, huang2024words}. Explicit capability representation is also particularly important. The automatically generated Robot Resumes in EMOS yield a substantial improvement over prompting without capability grounding, since without such representations LLMs may assign tasks to physically incapable robots. Finally, scalability remains a recurring limitation. Most surveyed systems show measurable degradation as robot count or context length grows, with few approaches validated beyond roughly ten robots. This points to context dilution and cascading pipeline errors as the primary bottlenecks for current systems.}

\subsubsection{Complex Task Decomposition}
 Task decomposition refers to scenarios where MRS must collaborate to complete one or more complex tasks that require careful planning and division of labor. In such cases, LLM can be leveraged to break down the overall task into smaller, manageable subtasks that align with the capabilities of each robot in the team. By designing effective prompts, LLMs can generate logical and actionable task decompositions, ensuring that the workload is distributed efficiently and that the robots cooperate seamlessly to achieve the overarching objective.

\begin{figure*}[ht]
    \centering
    \includegraphics[width=1\linewidth]{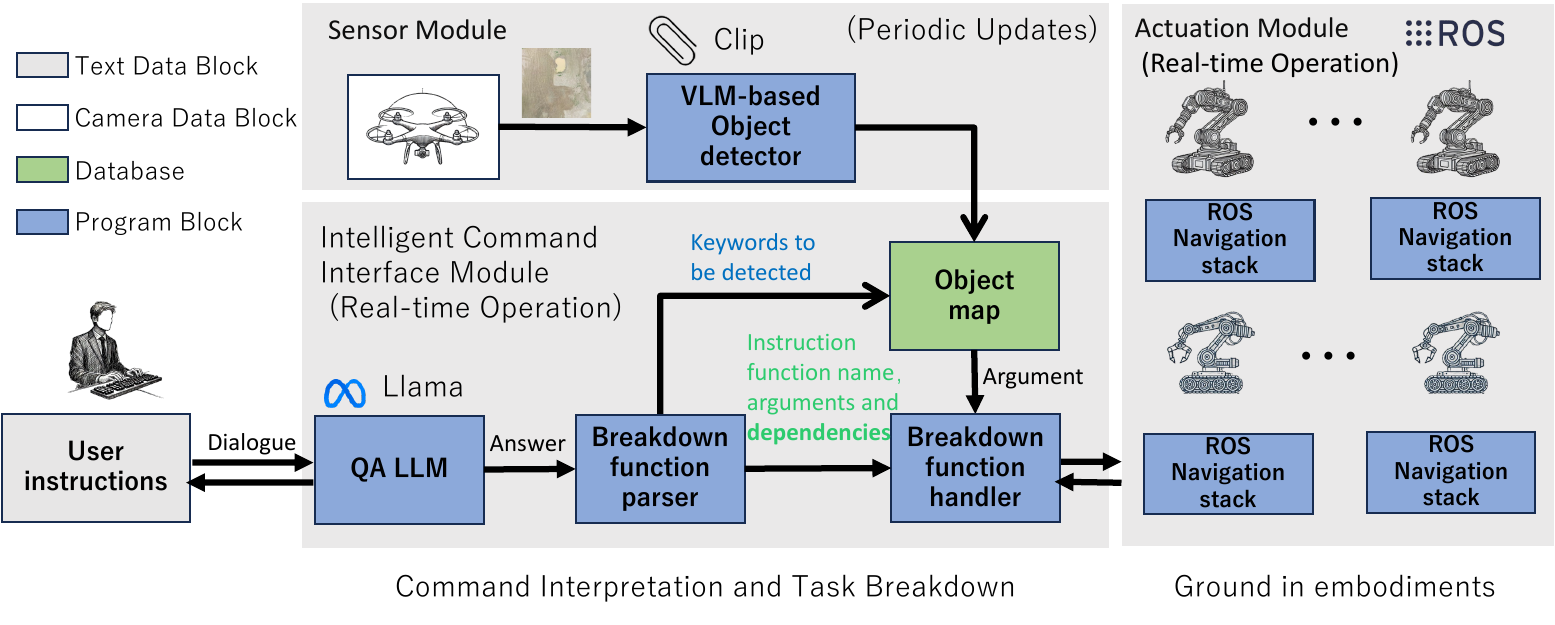}
    \caption{The framework of DART-LLM proposed by Wang \etal~\cite{wang_dart-llm_2024}. The system consists of three modules: Sensor Module, Intelligent Command Interface Module, and Actuation Module. The task decomposition process happens in the Intelligent Command Interface Module.}
    \label{fig:dart-llm}
\end{figure*}

Kannan \etal~\cite{kannan_smart-llm_2024} introduced \textbf{SMART-LLM}, a framework that utilizes LLMs to decompose high-level human instruction into subtasks and allocate them to heterogeneous robots based on their predefined skill sets. Unlike Chen \etal~\cite{chen_emos_2024}, where robot capabilities are inferred from their URDF code using LLMs, SMART-LLM adopts a more conventional approach by explicitly defining each robot's skill set for heterogeneous task allocation. The process involves decomposing instructions into sub-tasks, analyzing the required skills for each sub-task to form coalitions, and distributing robots accordingly to ensure efficient task execution.
Wang \etal~\cite{wang_dart-llm_2024} propose Dependency-Aware Multi-Robot Task Decomposition and Execution LLMs (\textbf{DART-LLM}), a system designed to address complex task dependencies and parallel execution problems for MRS, as shown in Fig.~\ref{fig:dart-llm}. The framework utilizes LLMs to parse high-level natural language instructions, decompose them into interconnected subtasks, and define their dependencies using a Directed Acyclic Graph (DAG). DART-LLM facilitates logical task allocation and coordination by establishing dependency-aware task sequences, enabling efficient collaboration among robots. Notably, the system demonstrates robustness even with smaller models, such as Llama 3.1 8B, while excelling in handling long-horizon and collaborative tasks. This capability enhances the intelligence and efficiency of MRS in managing complex composite problems. 
Xu \etal~\cite{xu_scaling_2024} proposed a two-step framework that leverages LLMs to translate complex natural language instructions into a hierarchical linear temporal logic (LTL) representation for MRS. In the first step, the LLM decomposes the instruction into a hierarchical task tree, capturing logical and temporal dependencies between subtasks to avoid errors in sequence. In the second step, a fine-tuned LLM translates each subtask into flat LTL formulas, enabling precise execution using off-the-shelf planners. This framework emphasizes the importance of temporal reasoning in decomposing complex instructions, ensuring accurate task allocation and execution for long-horizon and interdependent multi-robot tasks.
In contrast to the aforementioned approaches, Obata \etal~\cite{obata_lip-llm_2024} adopted a slightly different approach and proposed \textbf{LiP-LLM}, a framework integrating LLMs with linear programming for multi-robot task planning. Instead of providing end-to-end task allocation and execution, LiP-LLM utilizes LLMs to generate a skill set and a dependency graph that maps relationships and sequential constraints among tasks. The task allocation is then solved using linear programming to optimize task distribution among robots. This hybrid approach enhances task execution efficiency and success rates by combining LLMs' interpretative capabilities with the precision of optimization techniques. The results demonstrate the potential of integrating LLMs with traditional optimization techniques to improve the performance and coordination of MRS.
On the other hand, Liu \etal~\cite{liu_coherent_2024} proposed the \textbf{COHERENT} framework, which utilizes a Proposal-Execution-Feedback-Adjustment (PEFA) mechanism for task planning in heterogeneous MRS. The PEFA process involves a centralized task assigner LLM that decomposes high-level human instructions into subgoals and assigns them to individual robots. Each robot evaluates the assigned subgoal, determines its feasibility, and provides feedback to the task assigner, enabling dynamic adjustments and iterative refinements in the task plan. This process bears similarities to the robot discussion mechanism in the EMOS framework proposed by Chen \etal~\cite{chen_emos_2024}, where task decomposition and assignment leverage embodiment-aware reasoning based on robot resumes. However, COHERENT emphasizes a real-time, feedback-driven approach to handle task allocation and execution, making it particularly suited for dynamic and complex multi-robot environments.
Huang \etal~\cite{huang2025compositional} proposed LAN2CB (Language to Collective Behavior), a framework that converts natural language mission descriptions into executable multi-robot code. The system performs mission parsing, dependency analysis, and behavior tree construction, then generates task code using a library of goal generation, allocation, and motion primitives. LAN2CB supports dynamic replanning by updating the behavior tree and regenerating code when triggers occur, enabling robust adaptation in changing environments. Experiments on nine diverse scenarios in both simulation and real-world tests demonstrated high success rates and resilience to complex dependencies and environmental events.

Liang \etal~\cite{liang2025integrating} proposed a communication-based, feedback-driven framework that enhances multi-robot cooperation through a novel retrospective actor-critic paradigm. Their system comprises two large language models: $\text{LLM}_1$, which facilitates real-time collaborative discussion among robots to devise initial plans, and $\text{LLM}_2$, which serves as both an action critic and proposer. After executing a task or simulation, $\text{LLM}_2$ conducts a retrospective evaluation based on environmental feedback, critiques the earlier plans, and proposes improved alternatives. The framework leverages short-term and long-term memory to retain only the most recent retrospectives, thus maintaining contextual relevance while minimizing computational overhead. This approach demonstrates improved performance on RoCoBench tasks compared to baselines, highlighting the value of integrating LLM-based reflection into the decision-making pipeline for dynamic and uncertain multi-robot environments. Differently, Mandi \etal~\cite{mandi_roco_2024} proposed RoCo, a decentralized communication architecture for multi-robot collaboration, focusing on both high-level task planning and mid-level motion planning. In the RoCo framework, each robot is equipped with an LLM that engages in dialogue with other robots to discuss and refine task strategies. This dialogue process results in a proposed sub-task plan, which is validated by the environment for feasibility. If the plan fails (e.g., due to collisions or invalid configurations), feedback is incorporated into subsequent dialogues to improve the plan iteratively. Once validated, the sub-task plan generates goal configurations for robot arms, with a centralized motion planner computing collision-free trajectories. RoCo emphasizes flexibility and adaptability in multi-robot collaboration and has been evaluated using the RoCoBench benchmark, demonstrating robust performance across diverse task scenarios. This approach highlights the synergy between decentralized LLM-driven reasoning and centralized motion planning for complex, dynamic environments.
Peng \etal~\cite{peng2025automatic} introduced a knowledge-augmented automated MILP formulation framework for multi-robot task allocation and scheduling. Their approach leverages locally deployed LLMs, combined with a structured knowledge base, to extract spatiotemporal constraints from natural language and translate them into executable MILP code. The two-step pipeline first interprets fuzzy natural instructions into mathematical constraints and then generates optimized Gurobi-compatible code. Notably, the entire process is locally deployable, ensuring data privacy in sensitive industrial settings. Experimental results on aircraft skin manufacturing tasks demonstrate that the framework achieves high accuracy and efficiency in both model construction and code generation, highlighting the potential of integrating LLMs with symbolic optimization techniques for practical, privacy-sensitive MRS applications.
Cladera \etal~\cite{cladera2025air} demonstrated air-ground collaborative planning using LLM-based semantic reasoning and mapping for language-specified missions. Their decentralized system enables dynamic mission adaptation, as validated through extensive field experiments in both urban and rural settings, highlighting effective semantic-based collaboration and robust task execution.
Gupta \etal~\cite {gupta2025generalized} proposed a hierarchical tree-based mission planning approach that leverages LLMs for automated decomposition into manageable subtasks, explicitly tailored for heterogeneous robot teams. The proposed heuristic method efficiently schedules tasks, considering robot-specific constraints, and demonstrates flexibility and scalability across varied, complex missions.

\REV{A common two-stage design pattern emerges across these approaches. An LLM performs semantic parsing and structure extraction, such as subtask decomposition, DAG generation, LTL tree construction, or skill-graph synthesis. A downstream formal or combinatorial component, such as a temporal logic planner, linear programming solver, or DAG scheduler, then handles optimization and correctness enforcement~\cite{xu_scaling_2024, obata_lip-llm_2024, wang_dart-llm_2024}. This division reflects an implicit division of trust, in which LLMs are treated as reliable semantic translators but not as reliable optimizers. Such a choice is pragmatic, but it introduces a coupling problem when LLM-generated structures are plausible yet semantically wrong, causing solvers to compute optimal solutions to the wrong problem. Capability representation choices reveal a further trade-off. Predefined skill libraries~\cite{kannan_smart-llm_2024, obata_lip-llm_2024} reduce hallucination by constraining the LLM's output space but limit generalization to tasks outside the library, whereas open-ended prompt-embedded descriptions allow novel task handling at the cost of reliability. Systems that incorporate iterative feedback loops, including the PEFA mechanism of COHERENT~\cite{liu_coherent_2024}, the retrospective actor-critic of Liang \etal~\cite{liang2025integrating}, and the pre-execution dialogue of RoCo~\cite{mandi_roco_2024}, tend to outperform one-shot decomposition. However, each of these approaches shifts rather than eliminates the bottleneck. PEFA adds re-planning latency, the actor-critic roughly doubles LLM call overhead, and dialogue overhead grows rapidly with agent count.}

\subsection{Mid-Level Motion Planning}
Mid-level motion planning in MRS encompasses tasks such as navigation and path planning, where the focus lies on enabling robots to traverse or coordinate within an environment efficiently. These scenarios are more direct and practical than high-level applications, yet critical for the seamless operation of multi-robot teams. LLMs contribute significantly to this domain by leveraging their contextual understanding and learned patterns to generate robust and adaptive solutions. By interpreting environmental data and dynamically adapting to changes, LLMs enable robots to collaboratively plan paths, avoid obstacles, and optimize their motion in shared spaces. Integrating LLMs into mid-level motion planning enhances efficiency and resilience, making MRS more capable in dynamic and unpredictable environments. 

\begin{figure*}[ht]
    \centering
    \includegraphics[width=1\linewidth]{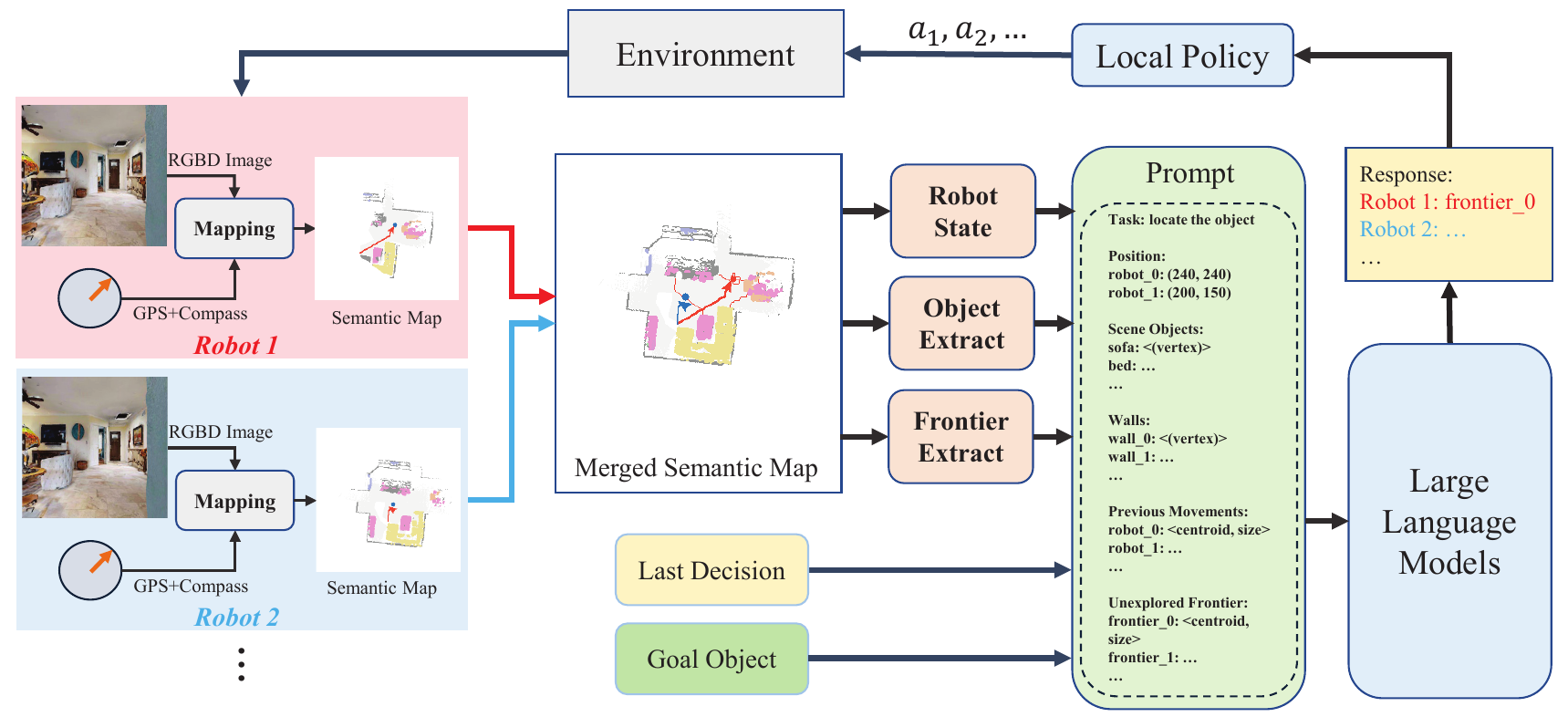}
    \caption{The framework of Co-NavGPT proposed by Yu \etal~\cite{yu_co-navgpt_2023}.}
    \label{fig:Co-NavGPT}
\end{figure*}

Yu \etal~\cite{yu_co-navgpt_2023} proposed \textbf{Co-NavGPT} framework to integrate LLMs as a global planner for multi-robot cooperative visual semantic navigation, as shown in Fig.~\ref{fig:Co-NavGPT}. Each robot captures RGB-D vision data, which is converted into semantic maps. These maps are merged and combined with the task instructions and robot states to construct prompts for the LLMs. The LLMs then assign unexplored frontiers to individual robots for efficient target exploration. By leveraging semantic representations, Co-NavGPT enhances environmental comprehension and guides collaborative exploration. 
In this framework, the LLMs are limited to allocating unexplored frontiers to each robot for navigation, serving primarily as a task allocation mechanism. 
Morad \etal~\cite{morad_language-conditioned_2024} took a step further and proposed a novel framework combining LLMs with offline reinforcement learning (RL) to address path-finding challenges in MRS. Their approach uses LLMs to translate natural-language commands into latent embeddings, which are then encoded with agent observations to produce state-task representations. Using offline RL, policies are trained on these representations to generate navigation strategies that understand and follow high-level natural language tasks. A key advantage of this framework is its ability to train policies entirely on real-world data without requiring simulators, ensuring direct applicability to physical robots. The integration of LLMs enhances the flexibility of task-instruction interpretation, while RL facilitates the generation of low-latency, reactive control policies, thereby enabling efficient multi-robot navigation.
Following this line, Godfrey \etal~\cite{godfrey_marlin_2024} developed \textbf{MARLIN} (Multi-Agent Reinforcement Learning Guided by Language-Based Inter-Robot Negotiation), a framework combining LLMs with Multi-Agent Proximal Policy Optimization (MAPPO) to enhance training efficiency and transparency in multi-robot navigation tasks. In MARLIN, robots equipped with LLMs engage in natural language negotiations to collaboratively generate task plans, which are then used to guide policy training. This hybrid approach dynamically switches between LLM-guided planning and standard MAPPO-based reinforcement learning, leveraging LLM reasoning to improve training speed and sample efficiency without sacrificing performance. Experimental results demonstrate that MARLIN achieves faster convergence and more consistent performance than conventional MARL approaches, with applications validated in both simulation and physical robot environments. This integration of negotiation-based planning highlights the potential of combining LLMs with MARL for scalable, explainable multi-robot coordination.
On the other hand, Garg \etal~\cite{garg_foundation_2024} used LLMs to resolve deadlocks in connected multi-robot navigation systems. In obstacle-laden environments, such systems can experience deadlocks that low-level control policies cannot resolve. To address this, the LLM selects a leader robot and plans waypoints for it to reach the target. The system reconfigures into a leader-follower formation, with a GNN-based low-level controller guiding the leader along the waypoints. 
Mahadevan \etal~\cite{mahadevan2025gamechat} directly addressed the challenge of multi-robot deadlocks by introducing \textbf{GameChat}, a role-based language-negotiation framework for collaborative decision-making. In this system, agents engage in natural-language dialogues to dynamically assign roles, such as ``leader'' or ``blocker'', that help resolve spatial conflicts, task interdependencies, and bottlenecks that typically lead to deadlocks. These roles emerge through decentralized negotiation, without requiring pre-defined hierarchies or centralized control. The framework demonstrates robust performance across a variety of collaborative tasks, including navigation and manipulation, in both simulation and real-world experiments. GameChat showcases how emergent role-based reasoning and interactive dialogue can effectively address deadlock resolution in complex multi-agent environments. Moreover, Wu \etal~\cite{wu_hierarchical_2024} proposed a mid-level action LLM that uses short-horizon inputs, such as tracking errors and control costs, to generate parameters for an optimization-based robot controller, enabling it to follow planned waypoints effectively.
While the aforementioned research primarily employs centralized systems where LLMs handle motion planning for all robots, Wu \etal~\cite{wu_camon_2024} developed a decentralized multi-robot navigation system for household tasks. In this framework, each robot is equipped with an LLM to enable communication and collaboration. Robots dynamically recognize and approach target objects distributed across multiple rooms. Leadership is dynamically assigned through a communication-triggered mechanism, with the leader robot issuing orders based on the global information it gathers. This flexible and decentralized leadership strategy enhances adaptability and efficiency in collaborative navigation scenarios.
Shen \etal~\cite{shen2025enhancing} proposed a framework of decentralized semantic planning based on multimodal chain-of-thought reasoning. The MCoCoNav architecture consists of three modules and two auxiliary components. The Perception module evaluates the exploration value of the current scene using multimodal CoT reasoning; the Judgment module determines whether a robot should explore a new frontier or revisit a history node by analyzing the global semantic map; the Decision module selects the next long-term navigation goal based on predicted frontier scores or historical node scores. If the robots are stuck or trapped in place, the system resets their goals to randomly sampled points on the global map to avoid deadlocks and ensure continued exploration. 
Ji \etal~\cite{ji2025collision} proposed a reinforcement learning framework that groups LLM planners with collision and reachability constraints, ensuring physically valid multi-robot motion plans. Evaluations in two BoxNet environments demonstrate that constraint-aware small-scale LLMs significantly outperform larger ungrounded models in task success and generalization, highlighting the effectiveness of explicit constraint integration for reliable multi-robot motion planning.
Rajvanshi \etal~\cite{rajvanshi2025sayconav} introduced \textbf{SayCoNav}, a LLM-based decentralized adaptive planning approach. By dynamically updating collaboration strategies among heterogeneous robots, SayCoNav significantly improves multi-object search efficiency and adapts to changing robot conditions during mission execution, as validated through extensive simulation studies.
Wang \etal~\cite{wang2025multi} presented \textbf{SAMALM}, a decentralized multi-agent LLM actor-critic framework that ensures socially aware robot navigation. Using parallel LLM actors and critics for verification and cooperative navigation, the system effectively balances local autonomy and global coordination, as validated across diverse multi-robot navigation scenarios.

\REV{The motion planning literature reveals three distinct LLM integration modes, namely a runtime global planner queried at each step~\cite{yu_co-navgpt_2023}, an offline encoder whose embeddings shape the policy input space without incurring runtime latency~\cite{morad_language-conditioned_2024}, and a sparse, event-triggered conflict resolver invoked only during deadlocks or spatial disputes~\cite{garg_foundation_2024, mahadevan2025gamechat}. The sparse event-triggered mode offers a useful practical observation. LLMs tolerate deliberative latency only at low frequencies, making episodic conflict resolution a better fit than continuous dense motion planning, since step-by-step LLM querying accumulates latency rapidly as environments grow. Decentralized approaches distribute inference cost across robots, but recent designs that add structured reasoning per agent, such as the chain-of-thought in MCoCoNav~\cite{shen2025enhancing} or the dual actor-critic passes of SAMALM~\cite{wang2025multi}, may be slower per agent than a centralized single-query system, showing that decentralization does not guarantee efficiency. Across most surveyed approaches, scalability to large fleets remains unvalidated. Most experiments involve small teams, typically fewer than ten robots, and shared structural weaknesses include partial observability, since LLMs receive aggregated map summaries rather than raw sensor streams, the absence of real-world multi-robot validation at scale, and fixed communication policies that degrade in dynamic environments.}

\subsection{Low-Level Action Generation}
Low-level action generation focuses on controlling robot motion or posture at the hardware level, translating high-level goals into precise control commands. These tasks are critical for ensuring smooth and efficient operations in dynamic environments. While LLMs offer contextual reasoning and adaptability, their performance in low-level tasks, which demand high precision and real-time responsiveness, is often limited compared to traditional control methods. Hybrid approaches that combine LLMs with optimization-based controllers or reinforcement learning show promise in leveraging LLMs' flexibility while maintaining the precision required for reliable robot actions.

\begin{figure}[tb!]
\centering
    \includegraphics[width=1\columnwidth, trim={1cm 4cm 0.6cm 5.6cm},clip]{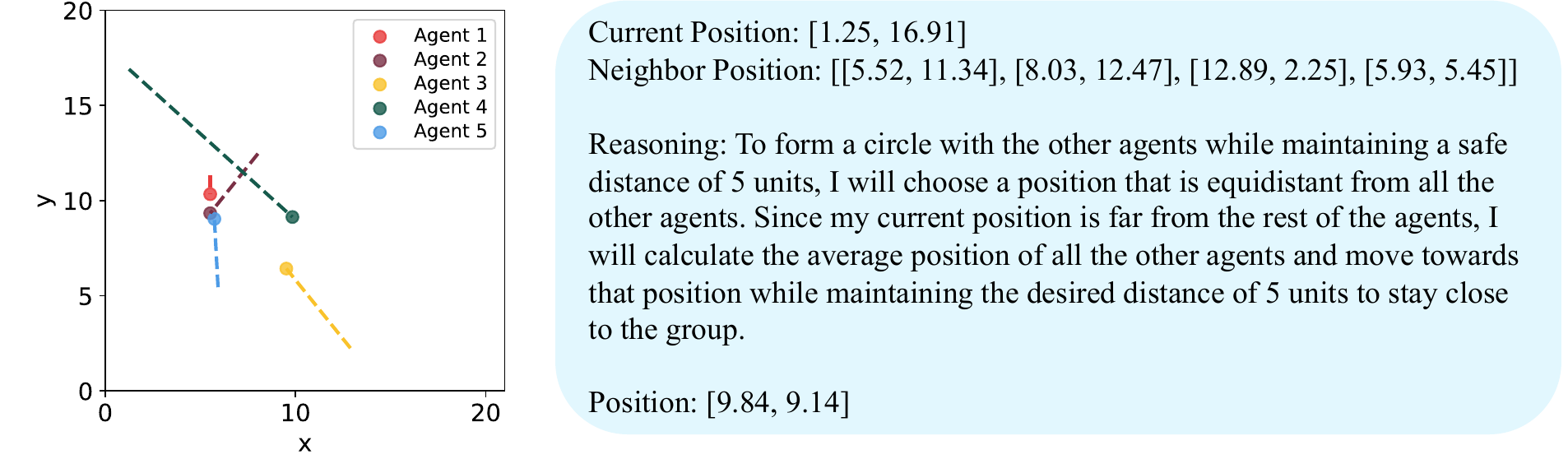}
    \includegraphics[width=1\columnwidth, trim={1cm 4cm 0.6cm 5.6cm},clip]{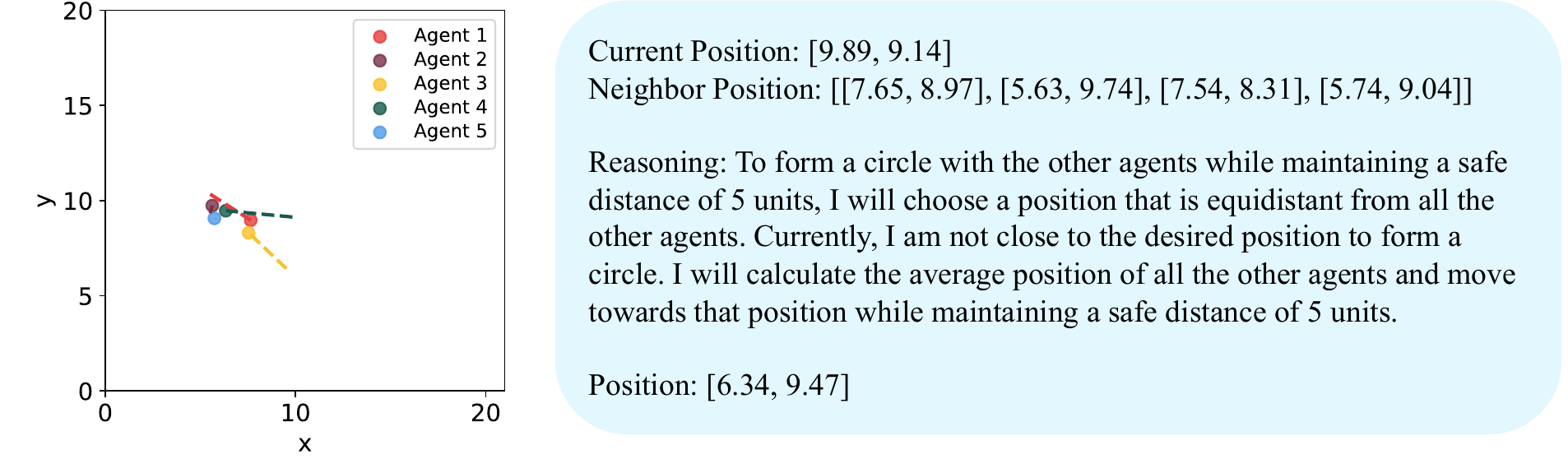}
    \includegraphics[width=1\columnwidth, trim={1cm 4cm 0.6cm 5.6cm},clip]{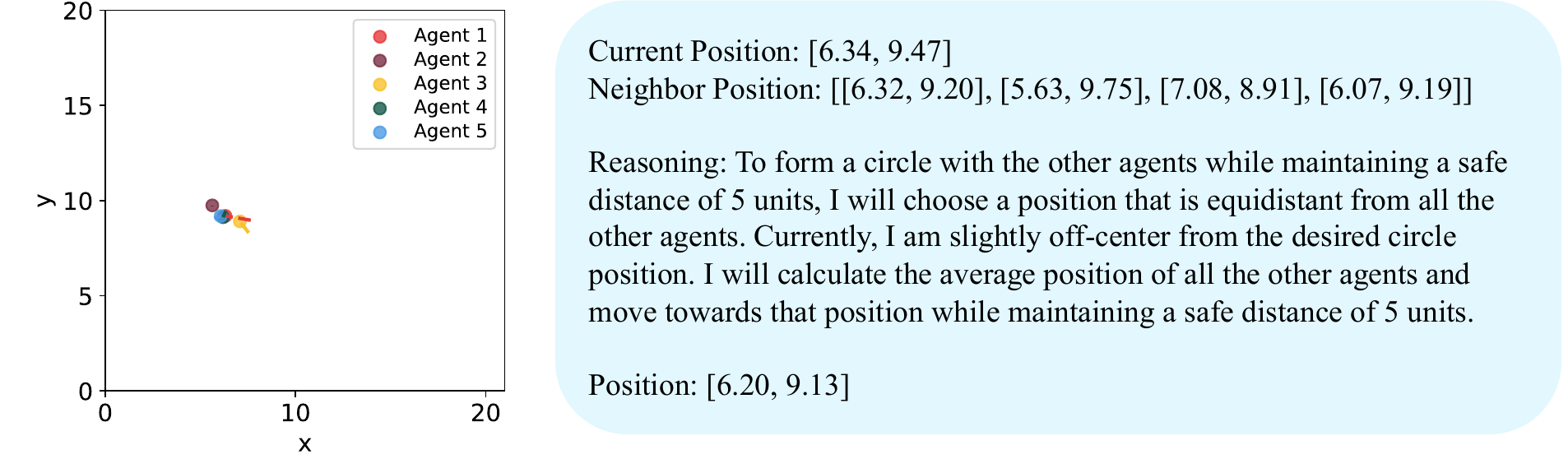}

    \vspace{2mm}
    \caption{Snapshots of five agents forming a circle~\cite{li_challenges_2024}. The desired distance between each agent is 5 units. The LLM's decision on Agent 4 is tracked.}
    \label{fig:Li}
\end{figure}

Chen \etal~\cite{chen_why_2024} leveraged LLMs to address the Multi-Agent Path Finding (MAPF) problem, where LLMs actively navigate robots by generating actions incrementally. Each step concludes with a high-level conflict checker to identify collisions with robots or obstacles. While effective in obstacle-free environments, LLMs struggle on maze-like maps due to limited reasoning capabilities, limited context length, and difficulty in understanding obstacle locations. Beyond path finding, most studies on using LLMs for action generation focus on formation control. For example, Venkatesh \etal~\cite{venkatesh_zerocap_2024} proposed a centralized architecture in which LLMs translate natural-language instructions into robotic configurations, enabling swarms to form specific patterns. Despite their strengths as centralized controllers, Li \etal~\cite{li_challenges_2024} highlighted the limitations of LLMs in decentralized systems. In a decentralized setup, each robot operates with its own LLM to achieve a desired formation through coordination with other robots. However, LLMs still face challenges in this task. In a test scenario shown in Fig.~\ref{fig:Li} where agents were tasked with forming a circle with a desired spacing of 5 units, the agent's LLM misinterpreted the instruction and directed the agent to move to the circle's center instead of the perimeter. This misunderstanding led the agent to exhibit consensus-based behavior rather than the intended flocking behavior, revealing the difficulties that LLMs face in distributed coordination.

In the field of formation control and planning, Strobel \etal~\cite{strobel_llm2swarm_2024} introduced \textbf{LLM2Swarm}, a system that integrates LLMs with robot swarms through two approaches: centralized controller synthesis and decentralized direct integration. In the centralized approach, LLMs are used to design and validate controllers prior to deployment, enabling efficient and adaptive behavior generation. In the decentralized approach, each robot maintains its own LLM instance, enabling localized reasoning, action generation, and collaboration, thereby enhancing flexibility in dynamic environments. The results highlight the potential of LLMs in swarm robotics, demonstrating their applicability to both centralized and decentralized control paradigms. Lykov \etal~\cite{lykov2024flockgpt} further showcased the potential of LLMs in swarm control with \textbf{FlockGPT}, a framework for orchestrating UAV flocks to achieve desired geometric formations. In this system, the LLM generates a signed distance function (SDF) to guide UAV motion relative to a target surface, while a dedicated control algorithm enforces practical constraints, such as collision avoidance. These studies underscore the versatility of LLMs in enhancing both centralized and decentralized swarm behaviors. Moreover, Xue \etal~\cite{xue2025formation} proposed a novel framework that leverages LLMs to achieve communication-efficient multi-robot formation control. The approach encodes formation goals, robot roles, and spatial arrangements into compact natural language instructions, which are then distributed to individual robots. Each agent decodes these instructions into actionable local goals. The framework demonstrates that LLM-generated descriptions can maintain formation integrity while significantly lowering communication bandwidth. Following this line, Ji \etal~\cite{ji2025genswarm} proposed \textbf{GenSwarm}, an end-to-end framework that enables automatic code-policy generation and deployment for multi-robot systems. The system interprets natural-language instructions to construct constraint-based skill graphs, and then hierarchically generates executable code policies using LLM agents. These policies are automatically reviewed, tested in simulation, and deployed to real-world robots via scalable hardware and software infrastructure. By supporting zero-shot generalization and human/VLM-in-the-loop feedback, GenSwarm offers a highly adaptive and interpretable solution for dynamic multi-robot tasks. Moreover, GenSwarm supports optional human-in-the-loop intervention, allowing users to provide natural-language feedback after observing execution outcomes. This feedback is interpreted by LLM agents to iteratively refine control policies, enabling intuitive human-in-the-loop policy optimization. \REV{Addressing the decentralized failure mode highlighted by Li \etal~\cite{li_challenges_2024}, Li and Zhou~\cite{li2025llm} proposed \textbf{LLM-Flock}, a decentralized multi-robot flocking framework that combines an influence-based consensus protocol in which robots iteratively adopt formation plans from more connected neighbors with a two-layer collision-avoidance strategy that pairs LLM-based urgency reasoning with a programmatic collision checker and re-prompting. Each robot's LLM is queried both at mission start to generate a candidate formation plan and at each motion step to reason about its next waypoint based on local neighbor positions. Validated across multiple closed-source and open-source LLMs in simulation and on physical Crazyflie drones, LLM-Flock demonstrates notable improvements in formation stability and convergence over prior decentralized LLM-based approaches.}

\REV{The most successful approaches in this subsection share a common architectural decision. The LLM generates a persistent, executable artifact such as a signed distance function~\cite{lykov2024flockgpt}, a Python control policy~\cite{ji2025genswarm, strobel_llm2swarm_2024}, or per-robot coordinate specifications~\cite{venkatesh_zerocap_2024}. This artifact is then executed by fast classical or physics-based controllers, instead of invoking the LLM at each control step. This separation is essential. When LLMs are instead queried per step for discrete actions~\cite{chen_why_2024}, performance degrades sharply in obstacle-laden environments or with more than a handful of agents, due to three compounding failures, namely imprecise spatial coordinate arithmetic, rapid context growth with agent count, and qualitative misinterpretation of geometric instructions. The centralized versus decentralized divide is a clear empirical contrast in this subsection. Centralized LLM-based formation controllers~\cite{venkatesh_zerocap_2024, lykov2024flockgpt, xue2025formation} consistently succeed by avoiding distributed spatial reasoning. Early fully decentralized architectures~\cite{li_challenges_2024} showed that individual LLM instances alone cannot reliably maintain inter-agent distance constraints, with agents in some tests drifting toward the group centroid rather than their intended perimeter positions. More recent work such as LLM-Flock~\cite{li2025llm} addresses this limitation by combining decentralized LLM-driven planning with an influence-based consensus protocol that iteratively refines local plans based on neighbor influence, demonstrating stable and adaptable flocking on both simulated and physical drones. The open challenge is closing the feedback loop. Current mechanisms such as the VLM video critique in GenSwarm~\cite{ji2025genswarm} and the nested LLM-PID loops in Xue \etal~\cite{xue2025formation} are promising but not yet reliable enough for dynamic real-world environments with unpredictable faults or rapidly changing task conditions.}

\subsection{Human Intervention}
In MRS, LLMs typically focus on executing tasks under human-provided instructions, emphasizing instruction interpretation and autonomous task completion. Once the instructions are delivered, human involvement is often minimized. However, emerging research explores scenarios that require continuous interaction between LLMs and humans, emphasizing cooperation, decision-making, or external observation throughout task execution. These studies highlight the potential for dynamic human intervention to address unexpected challenges, refine task strategies, or ensure safety in critical applications. By enabling iterative human-robot collaboration, such approaches enhance the adaptability and reliability of LLM-driven MRS.
The simplest form of human-robot interaction is demonstrated by Lakhnati \etal~\cite{lakhnati_exploring_2024}, where robots operate in a straightforward cycle: receiving a human command, executing the corresponding task, reporting the completion status, and awaiting the next instruction.
Building on this, Lykov \etal~\cite{lykov_llm-mars_2023} introduced the LLM-MARS framework, which enables humans to inquire about each robot's current state and task progress at any time. In this system, both response generation and task execution are handled by a single LLM, augmented with distinct LoRA adapters to improve efficiency.
Hunt \etal~\cite{hunt2024conversational} proposed an interactive approach that requires human approval before executing any plan generated through LLM-driven discussions. If the proposed plan is deemed unreasonable, the human supervisor can provide feedback, prompting the LLMs to refine their approach through further dialogue.
Ahn \etal~\cite{ahn_vader_2024} introduced the VADER system, further enhancing human involvement. When a robot encounters a task-related issue, it posts a request for assistance on the Human-Robot Fleet Orchestration Service (HRFS), a shared platform accessible to both human operators and robotic agents. Any agent or human can respond to the request, and once the issue is resolved, the robot resumes its task.
Li \etal~\cite{li2025hmcf} introduced HMCF, an approach that integrates LLM-driven multi-robot coordination with human oversight. Their framework mitigates hallucinations through human verification, significantly improving task success rates and providing robust zero-shot generalization across diverse robotic tasks.
These examples illustrate the varying degrees of human involvement in LLM-driven MRS, ranging from simple command execution to active collaboration and dynamic problem-solving.

\REV{Across these approaches, human intervention spans a spectrum from command-then-report cycles~\cite{lakhnati_exploring_2024}, to supervisory modes such as pre-execution approval~\cite{hunt2024conversational} or post-generation verification~\cite{li2025hmcf}, and on to on-demand modes where the robot initiates help requests when errors arise~\cite{ahn_vader_2024}. A recurring pattern is that human oversight is invoked primarily to mitigate LLM failure modes, rather than to enrich collective capability, in which verification catches hallucinated plans, approval gates prevent unsafe actions, and request channels handle out-of-distribution events. Two open gaps stand out across these systems. First, none treat the human as a peer negotiator whose preferences actively shape the plan, so the human's role remains reactive rather than collaborative. Second, none quantify the cognitive load of different intervention modes as team size scales, which is essential for understanding whether a single operator can oversee more than a handful of robots in real deployments.}

\begin{sidewaystable*}[!htbp]
\centering
\resizebox{1\textwidth}{!}{
\begin{tabular}{lccccccc}
\toprule
\textbf{Work} & \textbf{Communication} & \textbf{MRS Type} & \textbf{Modality} & \parbox[c]{1.5cm}{\centering \textbf{Human \\ Intervention}} & \textbf{Evaluation} & \textbf{Application} & \textbf{Model Type} \\
\midrule
\multicolumn{8}{l}{\textbf{High-Level Task Allocation and Planning}} \\
\midrule
Wu \etal~\cite{wu_hierarchical_2024}& Cent & Homo & T &  & Sim(ROS,RViz), Real & Target Tracking & GPT-4o, GPT-3.5 Turbo\\
Brienza \etal~\cite{brienza_llcoach_2024}& Cent & Homo & T, I &  & Sim(SimRobot) & Game & V: GPT-4 Turbo, L: GPT-3.5 Turbo \\
Lykov \etal~\cite{lykov_llm-mars_2023} & Cent & Homo & T, A & \checkmark & Real & General Purpose & Falcon \\
Ahn \etal~\cite{ahn_vader_2024} & Dec & Hetero & T, I & \checkmark & Real & General Purpose & V: CLIP, ViLD, PaLI, L: PaLM \\
Lakhnati \etal~\cite{lakhnati_exploring_2024} & Hier & Hetero & T, A & \checkmark & Sim(VR) & General Purpose & GPT-4\\
Chen \etal~\cite{chen_emos_2024} & Hier & Hetero & T &  & Sim & Household & GPT-4o\\
Lim \etal~\cite{lim2025dynamic} & Cent & Homo & T, I & & Real & Manufacturing & GPT-4o\\
Jiang \etal~\cite{jiangexploring} & Dec & Homo & T && Sim & General Purpose& GPT-4o, Gemini-2.0-Flash, DeepSeek-V3\\
Chen \etal~\cite{chen_scalable_2024} & Hier & Hetero & T &  & Sim(AI2THOR) & General Purpose & GPT-4, GPT-3.5 Turbo\\
Gupte \etal~\cite{gupte_rebel_2024} & Cent & Hetero & T &  & Sim(GAMA) & General Purpose & -\\
Huang \etal~\cite{huang2024words} & Cent & Homo & T & & Sim & General Purpose & GPT-4 Turbo\\
Kannan \etal~\cite{kannan_smart-llm_2024} & Cent & Homo & T &  & Sim(AI2THOR), Real & Household & GPT-4, GPT-3.5, Llama2, Claude3\\
Wang \etal~\cite{wang_dart-llm_2024} & Cent & Hetero & T, I &  & Sim & Construction & L: Llama-3.1, Claude 3.5, GPT-4o,GPT-3.5 Turbo, V: CLIP\\
Xu \etal~\cite{xu_scaling_2024} & Cent & Homo & T &  & Sim(AI2THOR+ALFRED), Real & Household & Minstral, GPT-4\\
Obata \etal~\cite{obata_lip-llm_2024} & Cent & Hetero & T &  & Sim(ROS) & General Purpose & GPT\\
Liu \etal~\cite{liu_coherent_2024} & Hier & Hetero & T &  & Sim(BEHAVIOR-1K), Real & General Purpose & GPT-4\\
Liang \etal~\cite{liang2025integrating} & Cent & Homo & T & & Sim(RoCoBench) & General Purpose & Llama3.1, Nemotron LLM\\
Mandi \etal~\cite{mandi_roco_2024} & Dec & Homo & T &  & Sim, Real & Household & GPT-4\\
Peng \etal~\cite{peng2025automatic} & Cent & Homo & T & & Sim & General Purpose & DeepSeek-R1-Distill-Qwen-32B, SFT-Qwen2.5-Coder-7B-Instruct\\
Yu \etal~\cite{yu_mhrc_2024} & Dec & Hetero & T &  & Sim(PyBullet) & Household & GPT-3.5 Turbo, GPT-4o, Llama3.1\\
Sueoka \etal~\cite{sueoka_adaptivity_nodate} & Dec & Hetero & T, I &  & Real & Construction & V: GPT-4v, L: GPT-4\\
Hunt \etal~\cite{hunt2024conversational} & Cent & Homo & T & \checkmark & Real & General Purpose & GPT-4\\
Yoshida \etal~\cite{yoshida_verification_nodate} & Dec & Hetero & T, I & & - & General Purpose & - \\
Wang \etal~\cite{wang_safe_2024} & Cent & Hetero & T & \checkmark & Sim(AI2THOR) & General Purpose & GPT-3.5, Llama-2, Llama-3\\
Guzman-Merino \etal~\cite{guzman-merino_llm_2024} & Cent & Hetero & T & & - & General Purpose & GPT-4o\\
Cladera \etal~\cite{cladera2025air}& Dec & Hetero & T, I &  & Sim, Real & Target Tracking & GPT-4o\\
Chen \etal~\cite{chen2025multi}& Cent & Homo & T &  & Sim & Formation & GPT-4o, GPT, DeepSeek-R1\\
Gupta \etal~\cite{gupta2025generalized}& Cent & Hetero & T &  & Sim & Field & GPT-4, Gemini, Llama\\
Wan \etal~\cite{wan2025toward}& Cent & Hetero & T, I &  & Sim, Real & Field & GPT-4, fine-tuned MultiPlan\\
Huang \etal~\cite{huang2025compositional} & Cent & Homo & T &  & Sim, Real & Formation & GPT-4.1\\

\midrule
\multicolumn{8}{l}{\textbf{Mid-Level Motion Planning}} \\
\midrule
Yu \etal~\cite{yu_co-navgpt_2023} & Cent & Homo & T &  & Sim(HM3D) & Household & GPT-3.5 Turbo\\
Morad \etal~\cite{morad_language-conditioned_2024} & Dec & Homo & T &  & Real & General Purpose & GTE-Based\\
Godfrey \etal~\cite{godfrey_marlin_2024} & Cent & Homo & T &  & Sim, Real & General Purpose & Llama-3.1\\
Garg \etal~\cite{garg_foundation_2024} & Cent & Homo & T, I &  & Sim, Real & General Purpose & V: Claude-3 Sonnet/Opus, GPT-4 Turbo, GPT-4o, L: GPT-4, GPT-3.5, Claude-2, Claude-3\\
Mahadevan \etal~\cite{mahadevan2025gamechat} & Dec & Homo & T & & Sim & Game & GPT-4o-mini\\
Wu \etal~\cite{wu_hierarchical_2024}& Cent & Homo & T &  & Sim(ROS, RViz), Real & Target Tracking & GPT-4o, GPT-3.5 Turbo\\
Wu \etal~\cite{wu_camon_2024} & Dec & Homo & T &  & Sim & Household & GPT-4o\\
Shen \etal~\cite{shen2025enhancing} & Dec & Homo & T, I & & Sim(HM3D, MP3D) & Target Tracking & GLM-4V-9B\\
Ji \etal~\cite{ji2025collision} & Dec & Homo & T &  & Sim & Warehouse & GPT-3.5, GPT-4\\
Rajvanshi \etal~\cite{rajvanshi2025sayconav} & Dec & Hetero & T &  & Sim & Household & GPT-4\\
Wang \etal~\cite{wang2025multi} & Dec & Homo & T &  & Sim & Household & GPT-4o\\

\midrule
\multicolumn{8}{l}{\textbf{Low-Level Action Generation}} \\
\midrule
Chen \etal~\cite{chen_why_2024} & Cent & Homo & T &  & Sim & General Purpose & GPT-4\\
Venkatesh \etal~\cite{venkatesh_zerocap_2024} & Cent & Homo & T, I &  & Sim(Pygame), Real & General Purpose & GPT-4, Llama-2, Claude-3 Opus\\
Li \etal~\cite{li_challenges_2024} & Dec & Homo & T &  & Sim & General Purpose & GPT-3.5 Turbo\\
Strobel \etal~\cite{strobel_llm2swarm_2024} & Dec & Homo & T &  & Sim(ARGoS), Real & General Purpose & GPT-3.5, GPT-4\\
Lykov \etal~\cite{lykov2024flockgpt} & Cent & Homo & T &  & Sim & Formation & GPT-4\\
Xue \etal~\cite{xue2025formation} & Cent & Homo & T &  & Sim & Formation & GPT-4 Turbo\\
Ji \etal~\cite{ji2025genswarm} & Cent & Homo & T, I & \checkmark & Sim(ROS) & Formation & o1-mini, GPT-4o, MetaGPT, CaP\\
Li and Zhou~\cite{li2025llm} & Dec & Homo & T &  & Sim, Real & Formation & GPT-5, Claude 4.5, Llama3.1-405b, Qwen-Max, DeepSeek-R1\\
Li \etal~\cite{li2025hmcf} & Dec & Hetero & T, I & \checkmark & Sim & Field & GPT-4\\

\bottomrule
\end{tabular}
}
\caption{Comparison of LLM-based MRS. Abbreviations: Communication: Cent (Centralized), Dec (Decentralized), Hier (Hierarchical); System: Homo (Homogeneous), Hetero (Heterogeneous); Modal: T (Text), I (Image/Video), A (Audio); Evaluation: Sim (Simulation), Real (Real-world experiments); Model Type: V (VLM), L (LLM).}

\label{tab:table1}

\end{sidewaystable*}

\section{Applications}\label{sec:5-application}
The integration of LLMs into MRS has enabled advancements across a variety of application domains, each with unique challenges and opportunities. These applications leverage LLMs' capabilities in understanding, task planning, and coordinating tasks, offering solutions ranging from indoor to outdoor scenarios. The adaptability of LLMs has driven innovation in tasks requiring precise navigation, task allocation, and dynamic decision-making, demonstrating their potential to address both structured and unstructured environments.

In this section, we categorize studies based on their application scenarios, focusing on two primary domains. First, the household domain highlights MRS addressing indoor challenges such as navigation, task decomposition, and object manipulation. These systems often emphasize collaboration among heterogeneous robots to execute intricate tasks, from identifying targets in multi-room settings to organizing household appliances. Second, applications in construction, formation, target tracking, and game illustrate the versatility of LLMs in specialized fields. These studies showcase MRS solving complex problems in outdoor or competitive environments, such as drone formations for search-and-rescue missions, robotic soccer strategies, and navigation in hazardous areas. Together, these domains underscore the growing impact of LLMs in advancing MRS capabilities across diverse real-world contexts.\\

\noindent \textbf{Household.} The household domain represents a significant focus in studies with well-defined application scenarios, addressing challenges such as navigation, task allocation, and task decomposition. For example, Wu \etal~\cite{wu_camon_2024} and Yu \etal~\cite{yu_co-navgpt_2023} investigated navigation and multi-target localization in complex indoor environments, such as identifying objects across multiple rooms, showcasing advancements in spatial awareness and adaptability. Furthermore, Mandi \etal~\cite{mandi_roco_2024}, Yu \etal~\cite{yu_mhrc_2024}, Kannan \etal~\cite{kannan_smart-llm_2024}, and Xu \etal~\cite{xu_scaling_2024} explored task decomposition and multi-robot collaboration to execute intricate tasks, such as preparing sandwiches or organizing dishwashers. Chen \etal~\cite{chen_scalable_2024} emphasized task allocation for heterogeneous MRS in multi-floor indoor settings, tackling coordination challenges in dynamic environments. Notably, Chen \etal~\cite{chen_emos_2024} proposed the EMOS framework, an embodiment-aware operating system that facilitates effective collaboration among heterogeneous robots through a novel ``robot resume'' approach, enabling robots to interpret their physical constraints from URDF files autonomously rather than relying on predefined roles. These studies address the temporal sequencing of subtasks while leveraging diverse robot capabilities, demonstrating the potential of MRS to solve complex, real-world problems in home environments.\\

\noindent\textbf{Other Applications: Construction, Formation, Target Tracking, and Games.}
Several studies focused on applications in open-world environments, highlighting the versatility and innovative potential of LLM-integrated robotic systems. For instance, Wang \etal~\cite{wang_dart-llm_2024} and Sueoka \etal~\cite{sueoka_adaptivity_nodate} explored the use of LLMs in orchestrating robotic systems for excavation and transportation tasks, showcasing their applicability in construction and complex terrain rescue operations. In drone formations applications, Lykov \etal~\cite{lykov2024flockgpt} emphasized the coordination and adaptability needed for outdoor tasks such as search-and-rescue missions and environmental monitoring. Similarly, Wu \etal~\cite{wu_hierarchical_2024} addressed open-world target tracking by integrating danger zone recognition, providing robust solutions for autonomous navigation in hazardous environments. Brienza \etal~\cite{brienza_llcoach_2024} introduced \textbf{LLCoach}, a framework for robotic soccer applications, where LLMs enhance strategic decision-making and team coordination. Collectively, these studies underscore the potential of LLM-driven MRS to tackle diverse and complex challenges across various domains. 

\section{LLMs, Simulations, and Benchmarks}\label{sec:6-benchmark}

\subsection{LLMs, VLMs, and VLAs}\label{sec:vlm-vla}
\subsubsection{Extending LLM Capabilities}
As discussed in Secs.~\ref{sec:3-comm}--\ref{sec:5-application}, LLMs form the backbone of many recent advances in MRS, enabling robots to follow natural language instructions, perform multi-step reasoning, allocate tasks, plan motions, and interact with human operators. Their strengths include flexible problem-solving across diverse scenarios, integration of symbolic or structured data, and the ability to generalize to new missions without retraining.

However, LLMs primarily operate on textual or symbolic representations of the environment. They rely on external perception modules to provide structured inputs such as maps, object lists, or state summaries. This separation between perception and reasoning can limit adaptability in visually complex or fast-changing environments where world states must be updated and integrated in real time.

To address these limitations, researchers have introduced multi-modal extensions, most notably VLMs and Vision-Language-Action models (VLAs)~\cite{zhong2025survey, o2024open}. VLMs incorporate visual information directly into the reasoning process, allowing robots to ground their decisions in raw sensory data~\cite{kim_openvla_2024, brohan2023rt}. VLAs take this one step further by integrating perception, reasoning, and action generation into a single pipeline, enabling faster, more reactive responses in dynamic settings~\cite{brohan2023rt}.

\subsubsection{VLMs vs. Traditional CV}
Traditional computer vision (CV) in MRS is typically task-specific, using separate modules for detection, segmentation, mapping, or action recognition. These components produce structured outputs that are passed to independent reasoning or low-level control systems. While this modularity can yield strong performance for narrow tasks, it often lacks the flexibility and adaptability needed for unstructured or dynamic missions.

VLMs bridge this gap by embedding visual features into a shared semantic space with language, enabling unified perception and reasoning. This allows recognition of previously unseen objects without retraining, switching between perception tasks by altering prompts, and direct connection between sensory input and high-level reasoning without intermediate, hand-coded logic. In MRS, these capabilities have been leveraged for perception-informed coordination, such as in Brienza \etal~\cite{brienza_llcoach_2024}, where a VLM coach transformed soccer match video into schematic scene descriptions for strategic task planning, and in the VADER framework~\cite{ahn_vader_2024}, where PaLI and CLIP grounded instructions in real-time affordance detection for replanning and recovery.

\subsubsection{VLA Models in MRS}
VLAs extend VLMs by coupling perception and reasoning directly with action generation~\cite{zhong2025survey}. Instead of outputting textual task plans, they can map visual observations and natural language commands to executable control sequences. This closed-loop capability shortens the cycle between sensing and actuation, allowing robots to respond faster to dynamic changes.

In an MRS setting, each VLA-equipped robot could combine the onboard camera feed, interpret the cooperative context, and directly issue motor commands for joint manipulation or navigation without routing through a separate task- or motion-planning layer. Although still relatively new in MRS research, VLAs are being explored in embodied AI for scenarios requiring fast perception–action loops, such as swarm coordination or hazard avoidance.

\subsubsection{Comparative Capabilities and Integration Patterns}
From an MRS perspective, LLMs, VLMs, and VLAs have complementary strengths. LLMs excel at long-horizon reasoning, symbolic task planning, and high-level negotiation between agents. VLMs add environmental grounding through perception, enabling multi-robot teams to share and reason about visual context. VLAs integrate perception and action for rapid, local decision-making.

Integration patterns in the literature generally follow three approaches: (1) \textit{visual grounding with language reasoning}, where a vision encoder produces scene features that an LLM uses for task planning and coordination~\cite{brienza_llcoach_2024, garg_foundation_2024}; (2) \textit{direct multi-modal prompting}, where visual and textual inputs are processed jointly for end-to-end reasoning~\cite{garg_foundation_2024, sueoka_adaptivity_nodate}; and (3) \textit{perception-to-action mapping}, where VLA-style architectures produce executable actions directly from multi-modal inputs, bypassing explicit textual reasoning.

Hybrid systems can combine these strengths. For example, a central LLM might manage mission-level strategy, VLMs on individual robots could process onboard sensor data for situational awareness, and VLAs could generate low-latency actuator commands in fast-changing conditions.

\subsection{Simulation Environments} 
 We have summarized the simulation platforms used in related works, highlighting their contributions to evaluating and advancing the field.
\textbf{AI2-THOR} has been adapted for MRS in \cite{chen_scalable_2024, kannan_smart-llm_2024, wang_safe_2024, xu_scaling_2024} to evaluate embodied AI agents operating in complex indoor environments \cite{ai2thor}. While originally designed for single-agent tasks such as object manipulation and scene understanding, recent research extends its use to MRS scenarios, including cooperative object retrieval, shared perception, and collaborative planning in constrained environments. The physics-enabled interactions allow researchers to test LLM-driven coordination strategies in dynamic and physically grounded environments, where multiple robots must navigate, manipulate objects, and resolve conflicts.
\textbf{PyBullet} is an open-source physics engine widely used for simulating robotic systems, including articulated manipulators, wheeled robots, and multi-robot interactions \cite{coumans2021}. It provides real-time physics simulations, supporting tasks like collision detection, rigid body dynamics, and reinforcement learning in robotics. In the context of MRS, PyBullet enables accurate modeling of decentralized collaboration, object manipulation, and dynamic environment interactions \cite{yu_mhrc_2024}.
\textbf{BEHAVIOR-1K}, utilized by Liu \etal~\cite{liu_coherent_2024}, serves as the foundation for the COHERENT framework, which focuses on large-scale, heterogeneous multi-robot collaboration \cite{li2023behavior}. This platform facilitates training and evaluation in complex household-like environments where different types of robots (e.g., manipulators, mobile bases) must coordinate to accomplish everyday tasks such as table setting, object handoff, and multi-step assembly processes. The benchmark ensures that LLM-enhanced systems can handle dynamic task dependencies and ambiguous role assignments.
The \textbf{Pygame} platform is a cross-platform Python module set designed for video game writing. Robots are modeled as point-mass entities, focusing on formation control, decentralized consensus algorithms, and motion coordination without obstacle avoidance. This platform is particularly useful for analyzing emergent behaviors in swarms, where LLM-based controllers guide self-organized formations through simple local interactions \cite{venkatesh_zerocap_2024}.
\textbf{Habitat-MAS}, an extension of Habitat, introduces explicit multi-agent communication for indoor navigation and exploration \cite{habitat19iccv, szot2021habitat, puig2023habitat3}. Unlike the single-agent focus of its predecessor, Habitat-MAS enables studies on cooperative search, SLAM, and inter-agent strategy adaptation, crucial for deploying multi-robot exploration teams in disaster response and service robotics \cite{chen_emos_2024}.
\textbf{ROS}-based simulation is a middleware framework widely used for MRS, enabling inter-robot communication~\cite{lampe2023robotkube}, decentralized control, and real-time data sharing. It provides essential tools for swarm coordination, collaborative mapping, and distributed task allocation. With built-in simulation environments like Gazebo and RViz, ROS allows researchers to develop and test MRS strategies for exploration, target tracking, and cooperative manipulation \cite{wu_hierarchical_2024}.
\textbf{VR} platforms introduce immersive simulations for human-robot collaboration and reinforcement learning. These environments are used to test human-in-the-loop control strategies for heterogeneous robot teams, such as coordinating robotic arms and mobile robots in warehouse logistics through natural language instructions \cite{lakhnati_exploring_2024}.
\textbf{GAMA} offers a multi-agent modeling environment suited for large-scale robot interactions \cite{gupte_rebel_2024}. It supports evaluations of distributed swarm intelligence, multi-agent task negotiation, and behavior adaptation in unstructured environments, making it ideal for testing decentralized LLM-driven controllers in logistics and autonomous fleet management.
\textbf{SimRobot}, utilized by Brienza \etal~\cite{brienza_llcoach_2024}, is specialized for multi-robot teamwork in robotic soccer. The LLCoach framework, trained using SimRobot, enhances robot coordination and strategic planning by processing match data and optimizing multi-robot role assignments dynamically.
\textbf{ARGoS}, chosen by Strobel \etal~\cite{strobel_llm2swarm_2024}, is a scalable platform for swarm robotics research. It enables controlled experiments on decentralized control mechanisms, including aggregate-then-disperse behaviors, leader election, and emergent self-organization. LLMs integrated into ARGoS are evaluated on their ability to generate adaptive communication protocols and handle task partitioning in dynamic environments.
These diverse platforms provide essential tools for evaluating LLM-driven MRS across different scales, from small collaborative teams to large autonomous swarms. By leveraging these environments, researchers refine multi-agent coordination, communication, and decision-making strategies, advancing the integration of LLMs in MRS for real-world applications.
\subsection{Benchmarks} 
Benchmarks are essential for evaluating LLM-driven MRS, providing standardized environments to measure coordination, adaptability, and performance across diverse scenarios. They enable consistent comparisons, helping identify strengths, limitations, and the effectiveness of MRS in real-world applications.
\textbf{RoCoBench}, introduced by Mandi \etal~\cite{mandi_roco_2024}, focuses on multi-robot manipulation tasks with human-in-the-loop support. The benchmark provides insights into multi-robot collaboration, particularly in tasks requiring shared manipulation, coordinated actions, and real-time adjustments to changing conditions. The benchmark provides detailed metrics on precision, task success rates, and robustness under unpredictable physical conditions, making it valuable for evaluating LLM-assisted multi-robot cooperation in human-shared workspaces.
\textbf{ALFRED}~\cite{shridhar2020alfred}, utilized in Xu \etal~\cite{xu_scaling_2024}, integrates language and vision benchmarks to test agents' ability to follow natural language instructions and execute multi-step tasks in household environments. Though originally focused on single-agent evaluations, ALFRED's framework can be extended to multi-robot task coordination, testing MRS on collaborative planning, sequential task execution, and efficient division of labor in domestic or service robotics applications.
\textbf{BOLAA}, proposed by Liu \etal~\cite{liu_bolaa_2023}, introduces a multi-agent orchestration benchmark specifically designed for LLM-augmented autonomous agents (LAAs). Unlike conventional evaluations that focus on individual agents, BOLAA assesses how LLMs manage multi-agent interactions, optimizing task distribution, decision-making, and real-time adaptability. This makes it a useful benchmark for LLM-driven MRS, where autonomous robots benefit from effective communication and collaboration to tackle complex, long-horizon objectives.
\textbf{COHERENT}-Benchmark, developed by Liu \etal~\cite{liu_coherent_2024}, is specifically designed for heterogeneous multi-robot collaboration in dynamic and realistic scenarios. Built on the BEHAVIOR-1K platform, this benchmark evaluates MRS across diverse environments, requiring coordinated task execution among robots with distinct capabilities, such as quadrotors for aerial mapping, robotic dogs for agile mobility, and robotic arms for precise manipulation. Key evaluation metrics include task allocation efficiency, inter-robot communication, and collaborative problem-solving, making it a comprehensive benchmark for testing LLM-driven coordination strategies in MRS.

\REV{Across the three robotics-oriented benchmarks, task success rate is reported universally. BOLAA, which targets software agents in WebShop and HotPotQA, instead reports domain-specific reward scores (attribute overlap on WebShop, F1 on HotPotQA) together with a retrieval recall metric on WebShop. Among the robotics benchmarks, RoCoBench and COHERENT-Benchmark additionally track the number of environment steps in successful runs, while ALFRED~\cite{shridhar2020alfred} adds path-length-weighted variants and goal-condition partial credit as standard evaluation metrics. Beyond these, LLM-specific measurements are largely absent. None of the four benchmarks reports hallucination rate, token consumption, API-call count, wall-clock inference latency, reasoning-chain quality (for example, step-level CoT validity or verifier-graded reasoning correctness), or robustness to perception noise as primary metrics, even though these factors feature prominently among the deployment concerns discussed in Sec.~\ref{sec:7-discussion}. The replan count in RoCoBench is the most direct proxy for LLM feedback utilization and is among the few LLM-sensitive quantities reported across the four benchmarks. Regarding multi-robot coverage, only RoCoBench and COHERENT-Benchmark are natively multi-robot. ALFRED is a single-agent household benchmark, and BOLAA orchestrates software agents rather than physical robots. A shared benchmark suite for LLM-based MRS should therefore cover communication overhead (tokens and messages per agent), inference latency and API-cost scaling curves as team size grows, hallucination and replanning frequency, coordination-quality metrics such as idle time and collision or conflict rate, and robustness tests under sensor noise and perturbations in heterogeneous teams. Establishing such metrics would enable meaningful comparisons of LLM-driven MRS beyond the current reliance on task success rate alone.}

\section{Challenges and Opportunities}\label{sec:7-discussion}
Despite the progress in integrating LLMs into MRS, significant challenges that limit their broader adoption and effectiveness remain. These challenges span areas such as reasoning capabilities, real-time performance, and adaptability to dynamic environments. Addressing these issues is critical to unlocking the full potential of LLMs in MRS. This section identifies key challenges the field faces and outlines promising opportunities for future research, offering a roadmap for enhancing the utility and robustness of LLM-driven MRS.

\subsection{Challenges}

\textbf{Insufficient Mathematical Capability.}
LLMs struggle with tasks that require precise numerical computation underlying mid-level motion planning and trajectory optimization. This limitation reduces their effectiveness in scenarios where quantitative accuracy is critical. Mirzadeh \etal~\cite{mirzadeh_gsm-symbolic_2024} performed a detailed comparison and investigation on the mathematical understanding and problem-solving ability of several state-of-the-art LLMs. Specifically, LLMs exhibit noticeable variance when responding to different variations of the same question, with performance declining significantly when only the numerical values are altered. Furthermore, their reasoning capabilities are fragile; they often mimic patterns observed in training data rather than performing genuine logical deduction. This fragility is exacerbated by an increase in the number of clauses within a question, even when the added clauses are irrelevant to the reasoning chain, leading to performance drops of up to 65\% in state-of-the-art models. These vulnerabilities present serious challenges for MRS, where precise calculations and robust reasoning are essential for collision-free trajectories, spatial planning, and efficient task execution. Addressing these limitations is critical for deploying LLMs reliably in mathematically intensive applications.\\

\noindent\textbf{Hallucination.}
LLMs are prone to generating content that appears plausible but lacks factual accuracy, a phenomenon known as hallucination. This issue is particularly concerning in MRS, where precise and reliable output is crucial for effective collaboration and operation. According to a comprehensive survey on hallucination in LLMs by Huang \etal~\cite{huang2023survey}, hallucination can be categorized into two main types: factuality hallucinations and faithfulness hallucinations. Factuality hallucinations involve discrepancies between generated content and verifiable real-world facts, leading to incorrect outputs. Faithfulness hallucinations occur when the generated content diverges from the user's instructions or the provided context, resulting in outputs that do not accurately reflect the intended information. In the context of MRS, such hallucinations can lead to misinterpretations, faulty decision-making, and coordination errors among robots, potentially compromising mission success and safety. Addressing these challenges requires developing methods to detect and mitigate hallucinations, ensuring that LLMs produce outputs that are both factually accurate and contextually appropriate.\\

\noindent\textbf{Multi-modal Integration Difficulties.}
While VLMs and VLAs extend the role of LLMs in MRS by grounding reasoning in perception and linking directly to action, they also introduce new challenges. Multi-modal inference often incurs higher latency due to the added cost of processing images, video, or sensor streams alongside text~\cite{li2024multimodal}. Domain shift remains a significant issue, as models trained on internet-scale vision-language corpora may not generalize reliably to the specialized sensing modalities of robots (e.g., LiDAR, thermal cameras, aerial views)~\cite{ma2024survey}. Moreover, fusing heterogeneous and asynchronous sensor data across multiple robots introduces synchronization and consistency problems, especially in decentralized settings~\cite{ma2024survey}. VLAs face the additional challenge of grounding action outputs across diverse robot morphologies, making it difficult to generalize learned behaviors across heterogeneous teams. Addressing these multi-modal integration difficulties is critical for ensuring that perception-grounded LLMs can operate robustly in real-world MRS deployments.\\

\noindent\textbf{Difficulties in Field Deployment.}
Current options for using LLMs include server-based models, which are usually closed-source, and open-source models that individuals can deploy locally. Examples of server-based models include OpenAI GPT~\cite{achiam2023gpt}, Anthropic Claude~\cite{claude}, and Google Gemini (formerly known as Bard)~\cite{googleGemini}, and open-sourced LLMs that individuals can run locally include Meta Llama~\cite{dubey2024llama}, Falcon~\cite{almazrouei2023falcon},  Alibaba Qwen~\cite{qwen2.5}, and DeepSeek V3~\cite{liu2024deepseek} and R1~\cite{guo2025deepseek}. The server-based models require a reliable internet connection to send inquiries and receive responses, thus making deploying the MRS with LLMs in remote locations unachievable, which is typical for field robot systems. Moreover, server-based LLMs heavily rely on the performance of the server, where a server outage can interrupt the systems built on LLMs entirely. This issue is especially vital for multi-robot teams as the LLM guides the inter-robot collaboration and decision-making. On the other hand, local models can avoid using the servers but require hardware onboard that is powerful enough to run LLMs locally. \\

\noindent\textbf{Relatively High Latency.}
Real-time information exchange and decision-making are critical for the effective operation of MRS in real-world scenarios. However, a notable challenge of using LLMs lies in their relatively high and variable response times, which can depend on model complexity, hardware capabilities, and server availability. For example, Chen \etal~\cite{chen_why_2024} reported that in a multi-agent path finding scenario utilizing GPT-4 from OpenAI, the response time per step ranged between 15 and 30 seconds, significantly impacting real-time feasibility. While local processing on more powerful hardware can reduce latency, this approach is costly and becomes less scalable as the number of robots increases. Addressing this challenge requires exploring optimized LLM architectures, efficient inference techniques, and scalable solutions that balance computational demands with real-time operational requirements.\\

\REV{\noindent\textbf{Deployment Readiness.}
Beyond the individual challenges discussed above, the collective deployment readiness of current LLM-based MRS systems can be assessed by comparing how representative works handle several real-world factors at once. Table~\ref{tab:deployment} summarizes how twelve representative systems spanning high-level task allocation, mid-level motion planning, and low-level action generation address LLM invocation frequency and cost, replanning, communication, observability, and evaluation scale. The columns jointly cover the deployment factors commonly invoked in field-robotics reviews. Feedback delays and latency are captured by the LLM invocation column, replanning frequency and error-triggered re-invocation by the replanning trigger column, sensing uncertainty and communication reliability by the communication and observability column, and end-to-end evaluation scale by the evaluation column. Several patterns emerge. First, wall-clock per-step latency is rarely reported. Only Lim \etal~\cite{lim2025dynamic} and SayCoNav~\cite{rajvanshi2025sayconav} provide concrete timings, while the remaining systems either move LLM inference out of the online decision loop (for example, GenSwarm~\cite{ji2025genswarm}, FlockGPT~\cite{lykov2024flockgpt}, and MARLIN~\cite{godfrey_marlin_2024}) or omit latency figures. Second, lossless communication is the dominant assumption. Only the SPINE-based air-ground system of Cladera \etal~\cite{cladera2025air, ravichandran2025spine} is explicitly designed for intermittent or opportunistic communication, whereas the other decentralized systems presume reliable neighbor message passing. Third, error recovery relies heavily on prompt-level re-invocation triggered by syntactic checks, validator failures, or discussion rounds. More substantive mechanisms, such as online odometry-drift correction or traversability-failure re-routing, appear only in SPINE~\cite{ravichandran2025spine}. Finally, physical validation remains small in scale. Across these systems, real-robot counts are typically in the range of two to eight, and no system in this table clearly exceeds roughly ten physical robots. Taken together, these observations suggest that LLM-based MRS has made strong progress at the algorithmic level, but deployment-grade robustness in the face of latency variance, lossy communication, and accumulated multi-robot uncertainty remains largely unaddressed.}\\

\newcolumntype{Y}[1]{>{\hsize=#1\hsize\raggedright\arraybackslash}X}
\begin{table*}[htbp]
\caption{\REV{Deployment factors across representative LLM-based MRS systems. ``Sim'' denotes simulation-only evaluation. ``Sim + Phys'' denotes simulation plus a small physical testbed. ``Field'' denotes outdoor or unstructured real-world deployment. ``n.r.'' denotes not reported. ``het.'' denotes heterogeneous. ``PEFA'' denotes the Proposal-Execution-Feedback-Adjustment cycle of COHERENT. ``SDF'' denotes Signed Distance Function.}}
\label{tab:deployment}
\centering
\footnotesize
\setlength{\tabcolsep}{4pt}
\renewcommand{\arraystretch}{1.2}
\begin{tabularx}{\textwidth}{@{}>{\raggedright\arraybackslash}p{2.2cm} Y{0.95} Y{0.85} Y{1.4} Y{0.8}@{}}
\toprule
\REV{System} & \REV{LLM invocation (reported latency/count)} & \REV{Replanning trigger} & \REV{Communication and observability} & \REV{Evaluation} \\
\midrule
\REV{HMAS-2~\cite{chen_scalable_2024}} & \REV{Per step (n.r.)} & \REV{Per-step consensus check and re-prompt} & \REV{Hybrid, text channel, full state via text} & \REV{Sim, up to 32 robots} \\
\REV{EMOS~\cite{chen_emos_2024}} & \REV{Pre-execution only (n.r.)} & \REV{Pre-execution discussion} & \REV{Hybrid, text channel, ideal semantic SLAM} & \REV{Sim, 2 robots/task (4 het. types)} \\
\REV{Lim \etal~\cite{lim2025dynamic}} & \REV{Per adaptation event (19.4\,s avg.)} & \REV{Event-triggered CCA validation} & \REV{Centralized, structured JSON state} & \REV{Sim + 2 manipulators} \\
\REV{COHERENT~\cite{liu_coherent_2024}} & \REV{Per PEFA cycle (n.r.)} & \REV{PEFA cycle on failure feedback} & \REV{Hybrid assigner-executor, text dialog, per-robot partial views} & \REV{Sim, up to 3 het. robot types} \\
\REV{RoCo~\cite{mandi_roco_2024}} & \REV{Per replan round (n.r., ${\sim}2.7$ on toy task)} & \REV{5-validator dialectic loop} & \REV{Decentralized dialog with central motion planner, asymmetric obs.} & \REV{Sim + HIL demo} \\
\REV{Co-NavGPT~\cite{yu_co-navgpt_2023}} & \REV{Per frontier event (n.r.)} & \REV{Frontier-event triggered} & \REV{Centralized VLM, per-robot local maps} & \REV{Sim + quadruped demo} \\
\REV{MARLIN~\cite{godfrey_marlin_2024}} & \REV{Training only (no LLM at deploy)} & \REV{Reactive policy at deploy} & \REV{Decentralized execution, partial obs.\ (POMDP)} & \REV{Sim + 2 TurtleBots} \\
\REV{SayCoNav~\cite{rajvanshi2025sayconav}} & \REV{Per step (${\sim}0.5$--$1.8$\,s wall-clock)} & \REV{Dynamic per-robot replanning} & \REV{Decentralized, text channel, per-robot local} & \REV{Sim, 1--3 agents, homo \& het.} \\
\REV{GenSwarm~\cite{ji2025genswarm}} & \REV{Once at code generation (${\sim}6$\,min)} & \REV{Regeneration on task change or feedback} & \REV{Hybrid, distributed local APIs at deployment} & \REV{Sim + Phys, custom multi-robot platform} \\
\REV{FlockGPT~\cite{lykov2024flockgpt}} & \REV{Once at SDF gen (n.r.)} & \REV{Manual re-prompt to update SDF} & \REV{Centralized, full state via motion capture} & \REV{Sim + 8 Crazyflies} \\
\REV{LLM-Flock~\cite{li2025llm}} & \REV{Per cycle (remote API)} & \REV{Influence-based consensus} & \REV{Decentralized, idealized neighbor broadcasts, per-robot local} & \REV{Sim + 5 Crazyflies} \\
\REV{Cladera \etal~\cite{cladera2025air} (SPINE-based)} & \REV{Per mission (2--4 API calls for short missions, ${\sim}37$--$41$ for large demos)} & \REV{Feedback-driven, active-perception re-query} & \REV{Intermittent comm.\ (by design), partial semantic graph} & \REV{Field, 1 UAV + 1 UGV} \\
\bottomrule
\end{tabularx}
\end{table*}

\REV{\noindent\textbf{Sim-to-Real Gap.}
While most LLM-based MRS studies validate their systems in simulation, relatively few evaluate on physical robot teams, and we are aware of few studies that report matched sim-versus-real success rate comparisons on identical LLM-driven tasks. The sim-to-real gap in LLM-based MRS differs from the traditional one in robot learning, as it primarily involves perception, representation, and latency rather than dynamics. Prompts and plans are typically tuned against idealized simulator states with perfect localization, noise-free sensing, and bounded actuation, all of which real hardware systematically violates. For example, Cladera \etal~\cite{cladera2025air} reported field failures caused by wind-blown leaves disrupting onboard LiDAR, false positives from open-vocabulary detectors, and communication loss, none of which surfaced in simulation. A second factor is the onboard inference barrier. Many real-world deployments route prompts to external servers or cloud APIs, as seen in GenSwarm~\cite{ji2025genswarm}, LLM-Flock~\cite{li2025llm}, and the air-ground system of Cladera \etal~\cite{cladera2025air}. This dependence on network connectivity and API latency becomes problematic in bandwidth-limited or infrastructure-free settings, where the zero-cost LLM inference assumed in simulation is no longer available. Potential mitigation directions include clamping LLM-emitted commands to physical limits, fine-tuning on deployment-matched data, and inserting consensus or human-in-the-loop critique layers to filter erroneous LLM outputs before they reach actuators. Closing this gap will require more realistic multi-robot simulators that model LLM latency variance, communication packet loss, and perceptual noise as core components of the testbed.}\\

\noindent\textbf{Lack of Benchmarks.}
Performance evaluation is essential for the new research on MRS with LLMs. However, existing benchmarking systems are primarily designed for indoor environments and household applications, which limits their applicability to the diverse and evolving scenarios where MRS operates. As current research often represents initial efforts to apply LLMs to MRS, performance comparisons typically focus on demonstrating feasibility by contrasting LLMs with traditional methods. While this approach is valuable for establishing a baseline, future advancements will likely yield significant performance and functionality improvements. A unified benchmarking framework tailored to multi-robot applications would provide researchers with consistent metrics to evaluate and quantify progress. Such a system would not only facilitate a clearer understanding of the impact of new research but also promote standardization and comparability across studies, accelerating innovation in this emerging field.

\subsection{Opportunities}

\textbf{Fine-tuning and RAG.}
Fine-tuning LLMs on domain-specific datasets and incorporating RAG techniques are promising avenues for improving their performance in multi-robot applications. Fine-tuning allows researchers to adapt pre-trained LLMs to specific tasks, enhancing their contextual understanding and reducing issues like hallucination. RAG complements this by integrating external knowledge retrieval mechanisms, enabling LLMs to access relevant information dynamically during runtime. Together, these techniques can significantly improve LLMs' accuracy, reliability, and adaptability in diverse and complex multi-robot scenarios.\\

\noindent\textbf{High-quality Task-specific Datasets.}
Creating high-quality and task-specific datasets is essential for advancing LLM capabilities in MRS. Leveraging more capable models, such as the latest LLMs, to generate synthetic datasets can accelerate the development of training materials tailored to specific tasks or environments. These datasets should include diverse scenarios, reasoning-focused labels, and context-specific knowledge to improve LLMs' problem-solving and decision-making capabilities. Task-specific datasets are particularly important for preparing MRS to operate in unstructured or open-world environments.\\

\noindent\textbf{Advanced Reasoning Techniques.}
Improving the reasoning capabilities of LLMs is critical for addressing their current limitations in logical and mathematical tasks. Techniques such as Chain of Thought (CoT) prompting~\cite{wei2022chain}, fine-tuning with explicit reasoning labels~\cite{zhang2023instruction}, integrating symbolic reasoning, and training with RL can enhance the ability of LLMs to handle complex multi-step problems. By advancing reasoning methods, LLMs can better support tasks that require precision and logical deduction, such as multi-robot path planning and coordination.\\

\noindent\textbf{Task-specific and Lightweight Models.}
While large-scale LLMs offer superior performance, they are often impractical for resource-constrained environments. Developing task-specific and lightweight models tailored for multi-robot applications can mitigate this issue~\cite{belcak2025small}. Models like SmolVLM, Moondream 2B, PaliGemma-2 3B, and Qwen2-VL 2B demonstrate how smaller architectures can reduce computational demands and latency while maintaining adequate performance for specific tasks. Model distillation is another approach to make small models more capable by distilling knowledge from a more capable LLM, like DeepSeek-R1-Distill-Qwen-1.5B, where the knowledge from DeepSeek R1 is distilled into a small Qwen2.5-Math-1.5B model~\cite{xu2024survey}. Balancing efficiency and effectiveness is key to enabling scalable deployments of LLMs in field robotics.\\

\noindent\textbf{Multi-modal and Embodied Extensions.}
Emerging research on multi-modal LLMs suggests promising opportunities for extending MRS capabilities. Video-language transformers, point cloud-language models~\cite{fu2021violet}, and spatio-temporal VLAs could enable richer reasoning over dynamic events~\cite{zheng2024tracevla}, 3D spatial layouts, and real-time cooperative interactions. These architectures would allow multi-robot teams to better interpret evolving environments and adapt their strategies accordingly. As multi-modal and embodied extensions mature, they could bridge the gap between high-level task planning and low-level action generation grounded in perception, unlocking more adaptive and autonomous multi-robot coordination~\cite{belcak2025small}.\\

\noindent\textbf{Expanding to Unstructured Environments.}
Most current applications and benchmarks focus on indoor or structured environments, leaving significant gaps in outdoor and unstructured scenarios. Research should prioritize expanding MRS capabilities to include operations in open-world contexts, such as agricultural fields, disaster zones, and remote exploration sites. Addressing the unique challenges of these environments, including variability, noise, and unpredictable dynamics, will broaden the applicability of LLM-enabled MRS.\\

\noindent\textbf{Latest More Capable LLMs.}
The continued development of state-of-the-art LLMs opens new possibilities for MRS. Models such as PaliGemma-2, Qwen-3, Claude 4 Sonnet/Opus, GPT o3 or o4-mini, GPT-5, and DeepSeek V3 and R1 offer enhanced reasoning, understanding, and multitasking capabilities~\cite{minaee2024large, xu2025towards}. Incorporating these advanced models into MRS research can accelerate progress by providing improved baseline performance and enabling innovative applications. Exploring their integration with robotics systems can further push the boundaries of what multi-robot teams can achieve.\\

\REV{\noindent\textbf{Realistic Multi-Robot Simulators for Sim-to-Real Transfer.}
Current multi-robot simulators model dynamics faithfully but typically treat LLM inference as zero-cost and instantaneous, and often assume lossless neighbor communication and noise-free perception. Future simulators should model these as core components, including LLM call latency and its variance, API cost scaling with team size, intermittent communication, and perceptual noise that propagates into VLM-based grounding. Combining such simulators with deployment-matched fine-tuning data and consensus or human-in-the-loop output validators offers a principled path to closing the LLM-MRS sim-to-real gap identified in the Challenges above.}\\

\REV{\noindent\textbf{Unified LLM-MRS Benchmarks and Metrics.}
A shared benchmark suite for LLM-based MRS would complement the task success rate that dominates current evaluations. Useful additions include communication overhead measured in tokens and messages per agent, inference-latency and API-cost scaling curves as team size grows, hallucination and replanning frequency, coordination-quality metrics such as idle time and conflict rate, and robustness tests under sensor noise and heterogeneous-team perturbations. Adopting such a suite would enable meaningful comparisons across LLM-driven MRS and identify which techniques actually reduce deployment risk.}
\section{Conclusion}\label{sec:8-conclusion}
This survey provides the first dedicated review of integrating LLMs into MRS, a topic at the intersection of robotics and artificial intelligence that is rapidly gaining prominence. Unlike general robotics or MAS, MRS poses unique challenges and opportunities due to their reliance on physical embodiment and real-world interaction. This paper highlights how LLMs can address these challenges, offering novel possibilities for collective intelligence and collaboration in MRS. \REV{Beyond cataloguing individual systems, we identify cross-paper design patterns at each planning level, assess collective deployment readiness through a comparative analysis of latency, replanning, communication, and evaluation scale, and discuss the sim-to-real gap and the absence of LLM-specific metrics in current benchmarks.}

We introduced a structured framework for understanding the role of LLMs in MRS, spanning high-level task allocation and planning, mid-level motion planning, low-level action generation, and human intervention. This framework reflects the diverse functionalities enabled by LLMs, including decomposing complex tasks, coordinating multi-robot multi-task scenarios, and facilitating seamless human-robot interaction. Additionally, we reviewed applications of MRS across various domains, from household tasks to construction, formation control, target tracking, and games and competitions, demonstrating the versatility and transformative potential of LLMs in these systems.

The significance of integrating LLMs into MRS lies in their ability to augment individual and collective intelligence, enabling robots to operate autonomously and collaboratively in increasingly complex environments. As LLMs continue to demonstrate their potential in everyday applications, their adoption in robotics promises to unlock new possibilities for innovation and efficiency in MRS.

Looking ahead, both immediate and long-term perspectives present exciting opportunities for research and development. In the near term, addressing challenges such as benchmarking, reasoning capabilities, and real-time performance will be critical for bridging the gap between lab-based simulations and real-world applications. Long-term prospects include leveraging LLMs to enable more complex missions, such as disaster response, space exploration, and large-scale autonomous operations, expanding the boundaries of what MRS can achieve.

We hope this survey will be a valuable resource for researchers, providing an overview of current progress, identifying gaps, and highlighting opportunities for future exploration. By advancing our understanding of LLMs in MRS, we aim to inspire innovation and foster collaboration across disciplines, accelerating the transition from theoretical studies to practical deployments that benefit society.

\section{Acknowledgment} 
We thank the authors of~\cite{liu_bolaa_2023, chen_scalable_2024, wang_dart-llm_2024, yu_co-navgpt_2023} for granting permission to include their original, unaltered figures in this paper.

\bibliography{reference}

\end{document}